\documentclass[twocolumn]{svjour3}      
\smartqed
\usepackage{natbib}
\usepackage{graphicx}
\usepackage{booktabs,multirow,array}
\usepackage{hyperref}
\usepackage{ragged2e}
\usepackage[ruled]{algorithm2e}
\usepackage{algorithmic}
\usepackage[misc]{ifsym} 
\usepackage{amsmath}
\usepackage{algorithmic}
\usepackage[caption=false,font=footnotesize]{subfig}
\usepackage {enumitem}
\setlist {nolistsep}
\usepackage{amssymb}
\usepackage{pifont}
\usepackage{xcolor}
\usepackage{makecell}

\begin{document}
\title{
Rethinking Portrait Matting with Privacy Preserving
\thanks{*S. Ma and J. Li are co-first authors and contribute equally to this work.\\
This work was supported by Australian Research Council Projects FL170100117 and IH180100002.}
}
\author{Sihan Ma$^{1*}$,
        Jizhizi Li$^{1*}$,
        Jing Zhang$^{1}$,
        He Zhang$^{2}$,
        Dacheng Tao$^{1}$
}

\institute{
Sihan Ma (sima7436@uni.sydney.edu.au) \\
Jizhizi Li (jili8515@uni.sydney.edu.au) \\
He Zhang (hezhan@adobe.com) \\
\Letter\  Jing Zhang (jing.zhang1@sydney.edu.au) \\
\Letter\  Dacheng Tao (dacheng.tao@gmail.com) \\
\\
$^{1}$ The University of Sydney, Sydney, Australia\\
$^{2}$ Adobe Inc., San Jose,  USA
}

\date{Received: date / Accepted: date}

\maketitle

%%%%%%%%%%%%%%%%%%%%%%%%%
%%% Section0. Abstract
%%%%%%%%%%%%%%%%%%%%%%%%%
\begin{abstract}
Recently, there has been an increasing concern about the privacy issue raised by identifiable information in machine learning. However, previous portrait matting methods were all based on identifiable images. To fill the gap, we present P3M-10k, which is the first large-scale anonymized benchmark for Privacy-Preserving Portrait Matting (P3M). P3M-10k consists of 10,421 high resolution face-blurred portrait images along with high-quality alpha mattes, which enables us to systematically evaluate both trimap-free and trimap-based matting methods and obtain some useful findings about model generalization ability under the privacy preserving training (PPT) setting. We also present a unified matting model dubbed P3M-Net that is compatible with both CNN and transformer backbones. To further mitigate the cross-domain performance gap issue under the PPT setting, we devise a simple yet effective Copy and Paste strategy (P3M-CP), which borrows facial information from public celebrity images and directs the network to reacquire the face context at both data and feature level. Extensive experiments on P3M-10k and public benchmarks demonstrate the superiority of P3M-Net over state-of-the-art methods and the effectiveness of P3M-CP in improving the cross-domain generalization ability, implying a great significance of P3M for future research and real-world applications. The dataset, code and models are available here\footnote{\url{https://github.com/ViTAE-Transformer/P3M-Net}}.
\end{abstract}

%%%%%%%%%%%%%%%%%%%%%%%%%
%%% Section1. Intro
%%%%%%%%%%%%%%%%%%%%%%%%%

\section{Introduction}

The success of deep learning in many computer vision and multimedia areas largely relies on large-scale of training data~\citep{zhang2020empowering}. However, for some tasks such as face recognition~\citep{masi2018deep}, human activity analysis~\citep{sun2019deep}, and portrait animation~\citep{9229197,Chen_2020_CVPR}, privacy concerns about the personally identifiable information in the datasets, e.g., face, gait, and voice, have attracted increasing attention recently. Unfortunately, how to alleviate the privacy concerns in data while not affecting the performance remains challenging and under-explored~\citep{yang2021study}. Specifically, portrait matting, which refers to estimating the accurate foregrounds from portrait images, suffers more from the privacy issue as most of the images contain identifiable faces in the previous matting datasets~\citep{dim,hatt,dapm}. Due to the population of the virtual video meeting in post COVID-19 pandemic era, this issue has received more and more concerns since portrait matting is a key technique in this multimedia application for changing virtual background. However, we found that all the previous portrait matting methods paid less attention to the privacy issue and adopted the intact identifiable portrait images for both training and evaluation, leaving privacy-preserving portrait matting (P3M) as an open problem.

\begin{figure}[t]
    \centering
    \includegraphics[width = 1\linewidth]{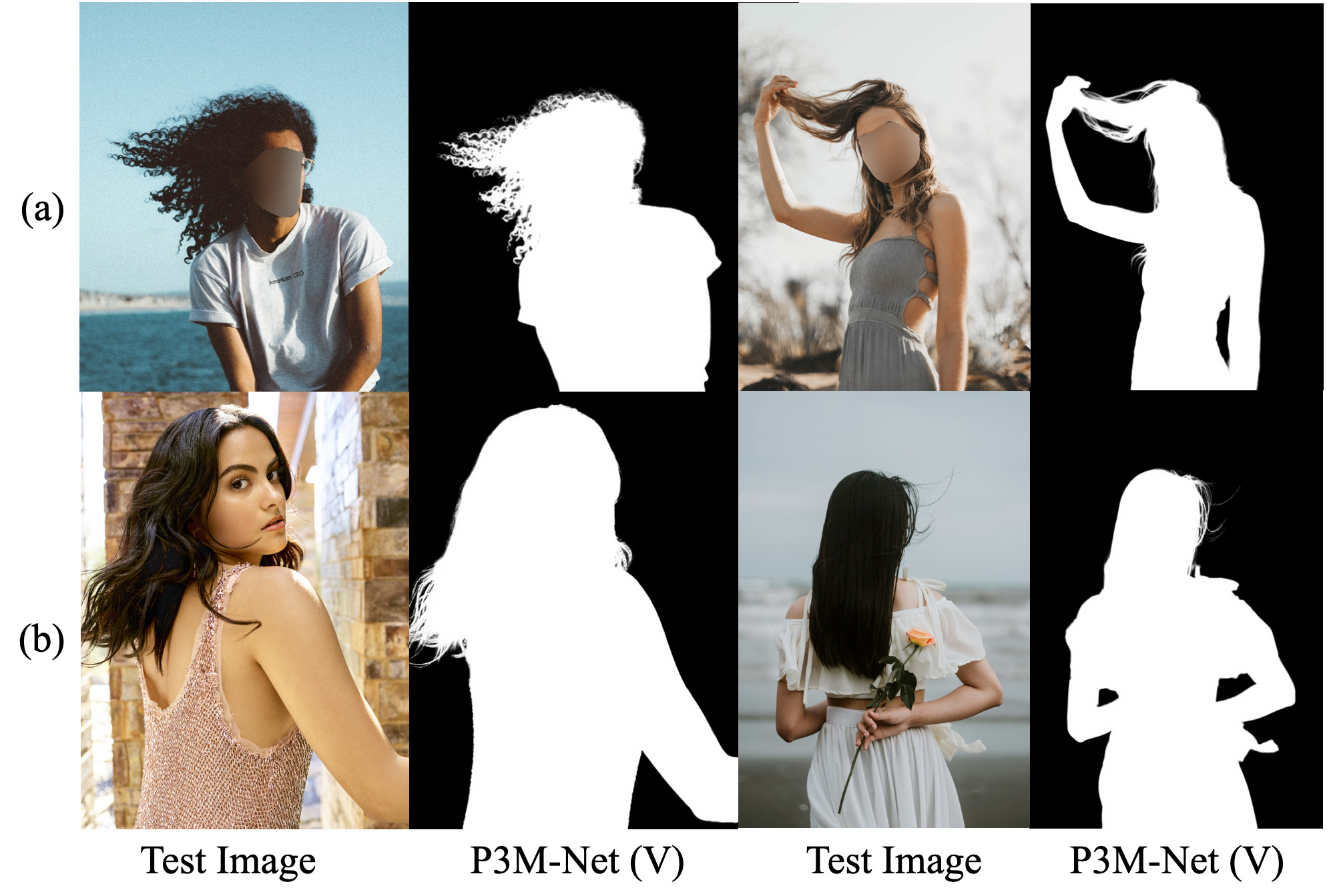}
    \caption{(a) Some anonymized portrait images from our P3M-500-P validation set. (b) Some non-anonymized celebrity or back portrait images from our P3M-500-NP validation set. We also provide the alpha mattes predicted by our P3M-Net variant P3M-Net (ViTAE), following the privacy-preserving training setting.}
    \label{fig:introduction}
\end{figure}

In this paper, we make the first attempt to address the privacy issue by setting up a new task P3M that requires training only on face-blurred portrait images (i.e., the Privacy-Preserving Training (PPT) setting) while testing on arbitrary images. We present the very first large-scale anonymized portrait matting benchmark P3M-10k consisting of 10,000 high-resolution face-blurred portrait images where we carefully collect and filter from a huge number of images with diverse foregrounds, backgrounds and postures, along with the carefully labeled high quality ground truth alpha mattes. It surpasses existing matting datasets~\citep{dapm,lf} in terms of diversity, volume and quality. Besides, we choose face obfuscation as the privacy protection technique to remove the identifiable face information while retaining fine details such as hairs. We split out 500 images from P3M-10k to serve as a face-blurred validation set, named P3M-500-P. Some examples are shown in Figure~\ref{fig:introduction}(a). Furthermore, to evaluate the generalization ability of matting models on non-privacy images when training on privacy-preserved images, we construct a validation set with 500 images without privacy concerns, named P3M-500-NP. All the images in P3M-500-NP are either frontal images of celebrities or profile/back images without any identifiable faces. Some examples are shown in Figure~\ref{fig:introduction}(b).

It can be observed from Figure~\ref{fig:introduction} that face obfuscation brings noticeable artefacts to the images which are not observed in normal portrait images. Then, one  common and interesting  question to explore is  that how will the proposed  PPT (Privacy-Preserving Training (PPT)) setting impact   existing SOTA matting models.   We notice that a contemporary work \citep{yang2021study} has shown empirical evidences that face obfuscation only has minor side impact on object detection and recognition models. However, the impact remains unclear in the context of portrait matting, where the pixel-wise alpha matte (a soft mask) with fine details is expected to be estimated from a high-resolution portrait image.

To address the above problem, we systematically evaluate both trimap-based \citep{dim,lu2019indices,li2020natural} and trimap-free matting methods \citep{shm,lf,modnet} on P3M-10k and provide our insights and analyses. Specifically, we found that for trimap-based matting, where the trimap is used as an auxiliary input, face obfuscation shows less impact on the matting models, i.e., a slight performance change of models following the PPT setting. As for trimap-free matting which involves two sub-tasks: foreground segmentation and detail matting, we found that the methods using a multi-task framework that explicitly model and jointly optimize both tasks~\citep{gfm,hatt} are able to obtain a good generalization ability on both face-blurred images and non-privacy ones at the same time. In contrast, matting methods that solve the problem in a  ``segmentation followed by matting'' manner \citep{shm,dapm} show a significant performance drop under the PPT setting. The main reason lies in the fact that  the segmentation errors led by face obfuscation may amplify error of the following matting model. Other methods that involve several stages of networks to progressively refine the alpha mattes from coarse to fine~\citep{shmc} seem to be less affected by face obfuscation but still encounter a performance drop due to the lack of explicit semantic guidance. Meanwhile, these methods require a tedious training process.

Based on the above observations, we propose a novel automatic portrait matting model P3M-Net, which is able to serve as a strong trimap-free matting baseline for the P3M task. Technically, we adopt a multi-task framework like proposed in~\citep{gfm, modnet} as our basic structure, which learns common visual features through a sharing encoder and task-aware features through a segmentation decoder and a matting decoder. To further improve the generalization abilities on P3M, we design a deep Bipartite-Feature Integration (dBFI) module to improve network's robustness ability to privacy-preserving training data by leveraging deep features with high-level semantics, a Shallow Bipartite-Feature Integration (sBFI) module to enhance network's ability to obtain fine details in the portrait images by extracting shallow features in the matting decoder, and a Tripartite-Feature Integration (TFI) module to promote the interaction between two decoders. We further design multiple variants of P3M-Net based on both CNN and vision transformer backbones and identify the difference of their generalization abilities. Extensive experiments on the P3M-10k benchmark provide useful empirical insights on the generalization abilities of different matting models under the PPT setting and demonstrate that P3M-Net along with its variants outperform all the previous trimap-free matting methods by a large margin.

Although our P3M-Net model achieves better performance than previous methods, there is still a performance gap when testing on face-blurred images and normal non-privacy portrait images, especially on face regions.  How to compensate for the lack of facial details in face-blurred training data to reduce the performance gap remains unresolved. To mitigate this problem, we devise an simple yet effective Copy and Paste strategy (P3M-CP) that can borrow facial information from publicly available celebrity images without privacy concerns and direct the network to reacquire the face context at both data and feature level. P3M-CP only brings a few additional computations during training, while enabling the matting model to process both face-blurred images and the non-privacy ones pretty well without extra effort during inference.

To sum up, the contributions of this paper are four-fold. \textbf{First}, to the best of our knowledge, we are the first to study the problem of privacy-preserving portrait matting and establish the largest privacy-preserving portrait matting dataset P3M-10k, which can serve as the benchmark for P3M. \textbf{Second}, we systematically investigate the impact of PPT setting and provide insights about the evaluation protocol, generalization ability and model design. \textbf{Third}, we propose a novel multi-task trimap-free matting model P3M-Net with three carefully designed interaction modules to enable privacy-insensitive semantic perception and details reserved matting. We then further devise multiple P3M-Net variants based on both CNN and vision transformer backbones to investigate their generalization abilities. \textbf{Fourth}, we devise a simple yet effective P3M-CP strategy that can improve the generalization ability of matting models for P3M under the PPT setting. Extensive experiments have demonstrated the value of our proposed methods, confirming P3M's ability to facilitate future research.

The remainder of the paper is organized as follows. Section~\ref{sec:lr} introduces the existing works related to matting, privacy issues in visual tasks, and vision transformer. In Section~\ref{sec:benchmark}, we systematically evaluate both trimap-based and trimap-free methods on P3M-10k, and analysize the impact of PPT setting. We introduce our P3M-Net model and its variants in Section~\ref{sec:p3mnet}, and present the copy and paste strategy P3M-CP in Section~\ref{sec:p3mcp}. Section~\ref{sec:experiments} manifests the subjective and objective experiment results of both P3M-Net and P3M-CP. Finally, we conclude the paper in Section~\ref{sec:conclusion}.

%%%%%%%%%%%%%%%%%%%%%%%%%
%%% Section2. RelatedWork
%%%%%%%%%%%%%%%%%%%%%%%%%

\section{Related Work}\label{sec:lr}

\subsection{Image Matting}
Image matting is a typical ill-posed problem to estimate the foreground, background, and alpha matte from a single image. Specifically, portrait matting refers to a specific image matting task where the input image is a portrait. From the perspective of input, image matting can be divided into two categories, i.e., trimap-based methods and trimap-free methods. Trimap-based matting methods use a user-defined trimap, i.e., a 3-class segmentation map, as an auxiliary input, which provides explicit guidance on the transition area. Previous methods include affinity-based methods~\citep{levin2007closed,aksoy2018semantic}, sampling-based methods~\citep{he2011global,shahrian2013improving}, and deep learning based methods~\citep{lu2019indices,hou2019context,Sun2021SemanticIM, Liu2021TripartiteIM}. Besides, there are other methods using different auxiliary inputs, e.g., a background image~\citep{backgroundmatting,backgroundmattingv2}, a coarse map~\citep{yu2021mask,dai2022boosting} or even language descriptions~\citep{rim}.

To enable automatic (portrait) image matting, recent works~\citep{shm,lf,hatt,gfm,aim} tried to estimate the alpha matte directly from a single image without using any auxiliary input, also known as trimap-free methods. For example, DAPM~\citep{dapm} and SHM~\citep{shm} tackled the task by separating it into two sequential stages, i.e., segmentation and matting. However, the semantic error produced in the first stage will mislead the matting stage and is difficult to be corrected. LF~\citep{lf} and SHMC~\citep{shmc} solved the problem by first generating coarse alpha matte and then refining it. Besides of the tedious training process, these methods suffer from ambiguous boundaries due to the lack of explicit semantic guidance. HATT~\citep{hatt} and GFM~\citep{gfm} proposed to model both the segmentation and matting tasks in a unified multi-task framework, where a sharing encoder was used to learn base visual features and two individual decoders are used to learn task-relevant features. However, HATT~\citep{hatt} lacks explicit supervision on the global guidance while GFM~\citep{gfm} and MODNet~\citep{modnet} lacks modeling the interactions between both tasks. By contrast, we propose a novel model named P3M-Net, which is also based on the multi-task framework but specifically focuses on modeling the interactions between encoders and decoders to better perform privacy-insensitive matting. Besides, several P3M-Net variants with CNN and vision transformer backbones have been designed and analysed.

\subsection{Privacy Issues in Visual Tasks}
There are two kinds of privacy issues in visual tasks, i.e., private data protection and private content protection in public academic datasets. For the former, there are concerns of information leak caused by insecure data transferring and membership inference attacks to the trained models~\citep{shokri2017membership,hisamoto2020membership,fredrikson2015model,carlini2019secret}. Privacy-preserving machine learning (PPML) aims to solve these problems based on homonorphic encryption~\citep{erkin2009privacy,yonetani2017privacy}, differential privacy~\citep{NEURIPS2020_fc4ddc15,pmlr-v161-he21a}, and federated learning~\citep{truex2019hybrid}.

For public academic datasets, there is no concern for information leak, thus PPML is no longer needed. But, there still exists privacy breach incurred by exposure of personally identifiable information, e.g., faces, addresses. It is a common problem in the benchmark datasets for many vision tasks, e.g., object recognition and semantic segmentation. Recently a contemporary work \citep{yang2021study} has shown empirical evidences that face obfuscation, as an effective data anonymization technique, only has minor side impact on object detection and recognition. However, since portrait matting requires to estimate a pixel-wise soft mask (alpha matte) for a high-resolution portrait image, the impact and difficulty remain unclear.

To explore the privacy-preserving portrait matting, a suitable face anonymization technique is necessary. A common method is to add empirical obfuscations~\citep{uittenbogaard2019privacy,caesar2020nuscenes,frome2009large,yang2021study}, such as blurring and mosaicing at certain regions. For the portrait matting task, we make the first attempt to construct a large-scale anonymized dataset named P3M-500-P and adopt face obfuscation as the privacy-preserving strategy. Specifically, we adopt multiple different face obfuscation methods, e.g., Gaussian blurring, mosaicing, zero masking, to construct different versions of P3M-500-P and validate the generalization ability of models trained with blurred images.

%%%%%% 3 matting datasets 
\subsection{Matting Datasets}
As shown in Table~\ref{tab:matting_datasets}, existing matting datasets either contain only a small number of high-quality images and annotations, or the images and annotations are in low-quality. For example, the online benchmark alphamatting~\citep{TUW-180666} only provides 27 high-resolution training images and 8 test images. None of them is portrait image. Composition-1k~\citep{dim}, the most commonly used dataset, contains 431 foregrounds for training and 20 foregrounds for testing. However, many of them are consecutive video frames, making it less diverse. GFM~\citep{gfm} provides 2,000 high-resolution natural images with alpha mattes, but they are all animal images. With respect to portrait image matting dataset, DAPM~\citep{dapm} provided a large dataset of 2,000 low-resolution portrait images with alpha mattes generated by KNN matting~\citep{chen2013knn} and closed form matting~\citep{levin2007closed}, whose quality is limited. Late fusion~\citep{lf} built a human image matting dataset by combining 228 portrait images from Internet and 211 human images in Composition-1k. Distinction-646~\citep{hatt} is a dataset containing 364 human images but with only foregrounds provided. There are some large-scale portrait datasets, e.g., SHM~\citep{shm}, SHMC~\citep{shmc}, and background matting~\citep{backgroundmatting}, which are unfortunately not public. Most importantly, no privacy preserving method is used to anonymize the images in the aforementioned datasets, making all the frontal faces exposed. By contrast, we establish the very first large-scale matting dataset with 10,000 high-resolution portrait images with high-quality alpha mattes, where all images are anonymized using face obfuscation.

%%% dataset comparison table
\begin{table*}[htb]
\begin{center}
\resizebox{\textwidth}{!}{
\begin{tabular}{c|ccccccc}
\hline
Dataset & Public & \#Images & \makecell[c]{\#Portrait\\Images} & \makecell[c]{Natural\\Images} & Resolution & Annotation & \makecell[c]{Privacy\\Preserving} \\
\hline
alphamatting.com~\citep{TUW-180666} & \checkmark & 35 & 0 & \checkmark & high & manually & \ding{55} \\
Composition-1k~\citep{dim} & \checkmark & 451 & 221 & \ding{55} & mixed & manually & \ding{55} \\
DAPM~\citep{dapm} & \checkmark & 2,000 & 2,000 & \checkmark & low & KNN, CF & \ding{55} \\
GFM~\citep{gfm} & \checkmark & 2,000 & 0 & \checkmark & high & manually & \ding{55}\\
Distinction-646~\citep{hatt} & \checkmark & 646 & 364 & \ding{55} & high & manually & \ding{55} \\
Late Fusion~\citep{lf} & \ding{55} & 258 & 258 & \ding{55} & high & manually & \ding{55} \\
SHM~\citep{shm} & \ding{55} & 35,513 & 35,513 & \ding{55} & - & manually & \ding{55} \\
SHMC~\citep{dai2022boosting} & \ding{55} & 9,449 & 9,449 & \ding{55} & - & manually & \ding{55} \\
P3M-10k (ours) & \checkmark & 9,921 & 9,921 & \checkmark & high & manually & face obfuscation \\
\hline
\end{tabular}}
\end{center}
\caption{Comparison of existing matting datasets. KNN and CF in column `Annotation' stand for obtaining the annotations of the dataset by KNN algorithm~\citep{chen2013knn} and closed-form algorithm~\citep{levin2007closed}, respectively.}
\label{tab:matting_datasets}
\end{table*}

%%%%%% 4 vision transformer
\subsection{Vision Transformer}
Transformer is a new deep neural structure that utilizes the self-attention mechanism to model the long-range dependency. It was first applied in the machine translation tasks in NLP, and shows great potential in vision tasks recently due to its superior representation capacity and flexible structure~\citep{dosovitskiy2020image,swin,vitae,zhang2022vitaev2}. ViT is the very first work~\citep{dosovitskiy2020image} to utilize pure transformer structures in image recognition and shows promising results. \citep{vitae} introduced the inductive bias into vision transformers to fully utilize the long-range modelling ability of self-attention and the locality and scale-invariance modelling ability of convolutions. Their ViTAE models achieved promising results in image recognition and many downstream tasks. Transformers also can handle multiple vision tasks at same time. Image Processing Transformer (IPT)~\citep{IPT} is proposed to process multiple low-level image tasks, including denoising, deraining, and super resolution. In matting, very few works have tried to explore and utilize the transformer related structure. In this work, we are the first to incorporate the vision transformers in our P3M-Net model to handle the global segmentation and detail matting tasks at same time, which manifests great generalization abilities.

\subsection{Comparison with the Conference Version}
Compared with our conference version~\citep{p3m_mm}, we extend the study by introducing three major improvements.
\textbf{First}, we rethink the problem of privacy-preserving portrait matting, and identify the performance gap of matting models between testing on face-blurred images and generalizing on normal non-privacy images. 
\textbf{Second}, we extend the P3M-Net model by proposing multiple variants based on both CNN and vision transformer backbones, and investigate the difference of their generalization abilities. This is the first time that vision transformers are adopted in trimap-free matting task. Extensive experiments validate that P3M-Net and its variants outperform all previous trimap-free matting methods on both face-blurred images and normal non-privacy images by a large margin, and even achieve comparable results with trimap-based matting methods.
\textbf{Third}, we introduce a simple yet effective P3M-CP strategy to mitigate the issue of absent face information during training under the PPT setting. It can borrow facial information from public celebrity images without privacy concerns and guide the network to re-acquire the face context at both data and feature level. Extensive experiments validate its effectiveness in improving the generalization ability of different matting models on normal non-privacy images.

%%%%%%%%%%%%%%%%%%%%%%%%%
%%% Section3. Benchmark
%%%%%%%%%%%%%%%%%%%%%%%%%

\begin{figure*}
    \centering
    \includegraphics[width=\linewidth]{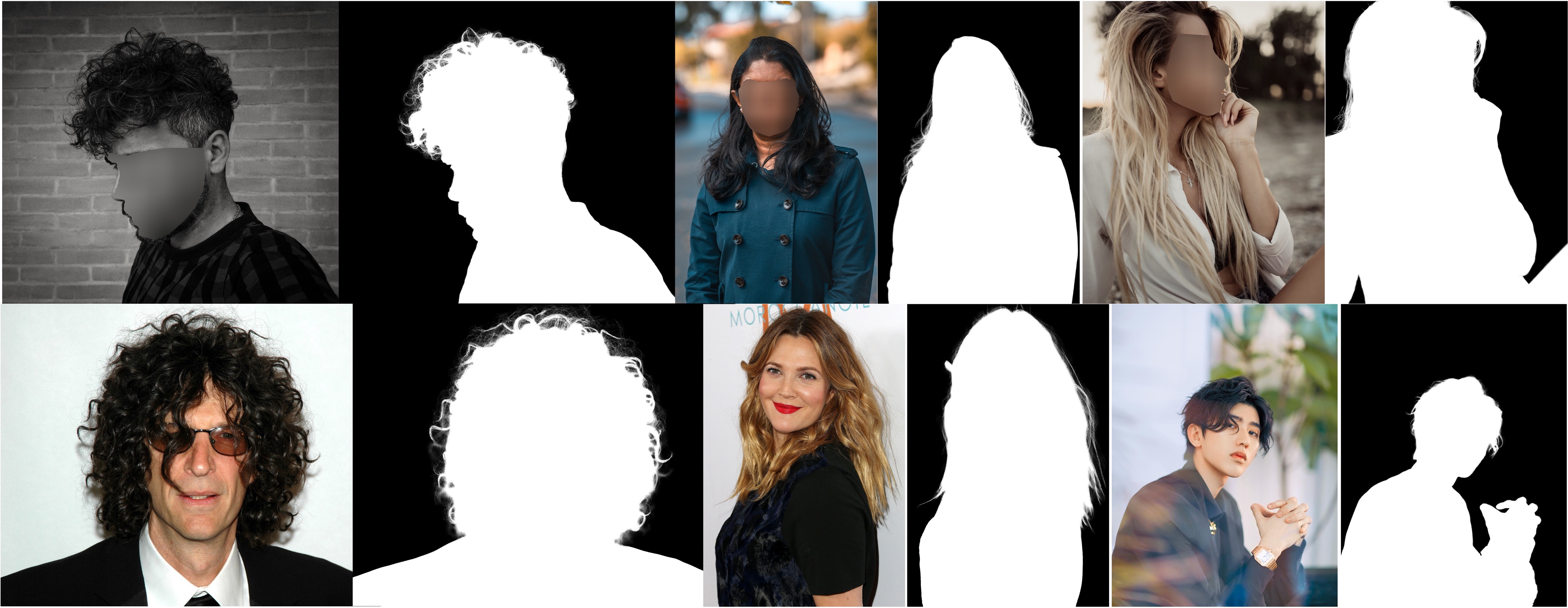}
    \caption{Top: samples from the P3M-10k training set and P3M-500-P validation set. Bottom: samples from the P3M-500-NP validation set.}
    \label{fig:data_p3m-10k}
\end{figure*}

\section{A Benchmark for P3M and Beyond}\label{sec:benchmark}
Privacy-preserving portrait matting is an important and meaningful topic due to the increasing privacy concerns. In this section, we first clearly define this new setting, then establish a large-scale anonymized portrait matting dataset P3M-10k to serve as the benchmark for P3M. A systematic evaluation of the existing trimap-based and trimap-free matting methods on P3M-10k is conducted to investigate the impact of the privacy-preserving training setting on different matting models. We then gain some useful insights in terms of the evaluation protocol, generalization ability, and model design.

\subsection{PPT setting}

Due to the privacy concern, we propose the \textbf{p}rivacy-\textbf{p}reserving \textbf{t}raining (PPT) setting in portrait matting, i.e., training only on privacy-preserved images (e.g., processed by face obfuscation) and testing on arbitrary images with or without privacy content. As an initial step towards privacy-preserving portrait matting problem, we only define the identifiable faces in frontal and some profile portrait images as the private content in this work. Intuitively, PPT setting is challenging since face obfuscation brings noticeable artefacts to the images which are not observed in normal portrait images, i.e., there is a clear domain gap between face-blurred images and normal images. How to eliminate the side impact of PPT setting and make model generalize well on normal images remains challenging but is of great significance for privacy-sensitive applications. 

\begin{figure*}
    \centering
    \includegraphics[width=\linewidth]{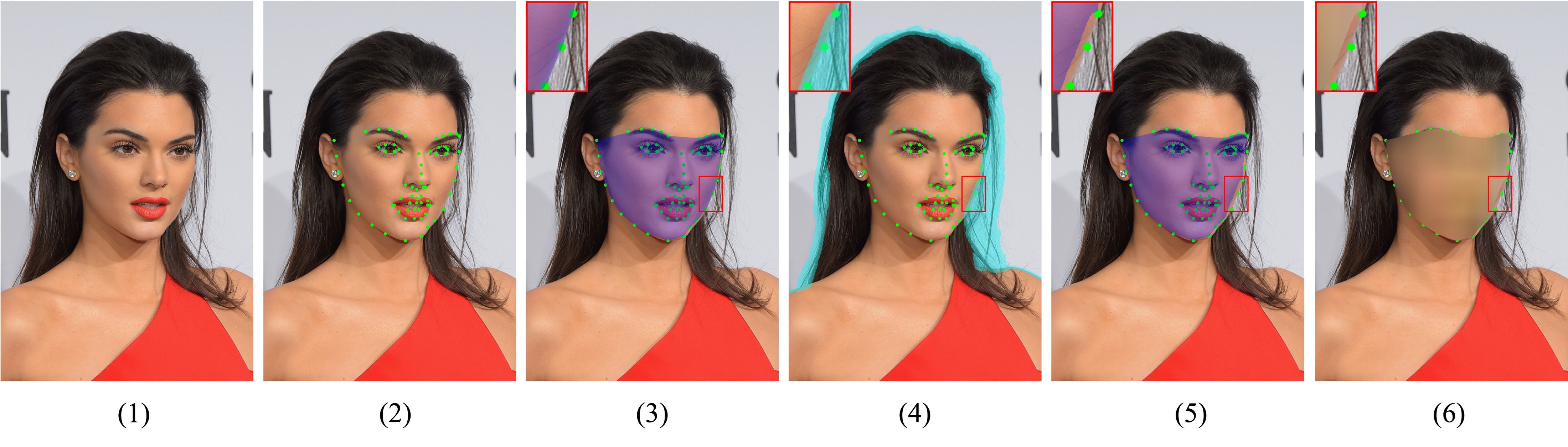}
    \caption{Illustration of the face blurring process. (1) Original image. (2) Generate facial landmarks (green dots). (3) Generate private area (purple mask). (4) Generate transition area (light blue mask). (5) Adjust private area, by excluding transition area. (6) Generate final blurred images (landmarks are only for reference).}
    \label{fig:face_obfuscation}
\end{figure*}

\subsection{P3M-10k dataset}

To answer the above question and provide a solid testbed for P3M, we establish the very first large-scale privacy-preserving portrait matting benchmark named P3M-10k. It contains 10,000 anonymized high-resolution portrait images by face obfuscation along with high-quality ground truth alpha mattes. Specifically, we carefully collect, filter and annotate about 10,000 high-resolution images from the websites with open licenses\footnote{\url{https://unsplash.com/} and \url{https://www.pexels.com/}}. We ensure all the images are not duplicate and have at least 1080 pixels on both two directions to ensure they are all high-resolution. We also check on each image to make sure it contains a salient and clear person.

As for the privacy-preserving method, we propose to use blurring to obfuscate the identifiable faces. Instead of using a face detector to obtain the bounding box of face and blurring it accordingly as in \citep{yang2021study}, we adopt a facial landmark detector to obtain the face mask. It is because different from the classification and detection tasks in \citep{yang2021study}, which may not sensitive to the blurry boundaries, portrait matting requires to estimate the foreground alpha matte with clear boundaries, including the transition areas of face such as cheek and hair. As shown in Figure~\ref{fig:face_obfuscation}, after obtaining the landmarks, a pixel-level face mask is automatically generated along the cheek and eyebrow landmarks in step (3). Then, we exclude the transition area shown in step (4) and generate an adjusted face mask at step (5). Finally, we use Gaussian blur to obfuscate the identifiable faces in the mask and the final result is shown in step (6). Note that for those images with failure landmark detection, we manually annotate the face mask.

Eventually, there are 9,921 images with face obfuscation remained. We then split them as 9,421 training images and 500 images noted as P3M-500-P to evaluate models' performance on face-blurred images in P3M. In addition, to evaluate models' generalization ability on non-privacy images in P3M, we further collect and annotate another 500 public celebrity images from the Internet without face obfuscation to form P3M-500-NP. Some examples of the training set and two validation sets are shown in Figure~\ref{fig:data_p3m-10k}.

Our P3M-10k outperforms existing matting datasets in terms of dataset volume, image diversity, privacy preserving, and providing natural images instead of composite ones. The diversity is not only shown in foreground, e.g., half and full body, frontal, profile, and back portrait, different genders, races, and ages, $etc$., but also in background, i.e., images in P3M-10k are captured in different indoor and outdoor environments with various illumination conditions. Some examples are shown in Figure~\ref{fig:data_p3m-10k}. In addition, we argue that large volume and high diversity of P3M-10k enable models to train on the natural images without the need of image composition using low-resolution background images, which is a common practice in previous works \citep{dim, hatt}, where they use composition to increase data diversity due to the small scale dataset volume but will bring in obvious composition artefacts due to the discrepancy of foreground and background images in noise, resolution, and illumination. The composition artefacts may have a side impact on the generalization ability of matting models as shown in \citep{gfm}. By contrast, the background in P3M-10k are compatible with the foreground since they are captured from the same scene.

\begin{table*}[htb]
\begin{center}
\resizebox{0.9\textwidth}{!}{
\begin{tabular}{c|c|cccccccc}
\hline
Val set & Metric & Closed & IFM & KNN& Compre & Robust& Learning & Global & Shared\\
\hline
\multirow{3}{*}{B} & SAD & 9.5750 & 10.887 & 15.378 & 8.3208 & 9.3321 & 10.248 & 9.6157 & 10.553 \\
 & MSE & 0.0214 & 0.0326 & 0.0511 & 0.0194 & 0.0214 & 0.0238 & 0.0242 & 0.0285 \\
 & MAD & 0.0693 & 0.0760 & 0.1087 & 0.0602 & 0.0674 & 0.0737 & 0.0708 & 0.0774 \\
 \hline
\multirow{3}{*}{N} & SAD & 9.4812 & 10.793 & 15.366 & 8.2295 & 9.2486 & 10.151 & 9.4908 & 10.386\\
 & MSE & 0.0210 & 0.0318 & 0.0506 & 0.0191 & 0.0211 & 0.0236 & 0.0236 & 0.0277\\
 & MAD & 0.0686 & 0.0748 & 0.1078 & 0.0595 & 0.0668 & 0.0729 & 0.0698 & 0.0760\\
 \hline
\end{tabular}}
\end{center}
\caption{Results of trimap-based traditional methods on the blurred images (``B'') and normal images (``N'') in P3M-500-P.}
\label{tab:benchmark_trimap_based_traditional}
\end{table*}

\begin{table*}[htb]
\begin{center}
\resizebox{0.9\textwidth}{!}{
\begin{tabular}{l|ccc|ccc|ccc}
\hline
Setting & \multicolumn{3}{c}{N:N} & \multicolumn{3}{|c}{B:N} & \multicolumn{3}{|c}{B:B} \\
\hline
Method & SAD & MSE  & MAD & SAD & MSE & MAD & SAD & MSE & MAD\\
\hline
DIM  & 4.7941 & 0.0116 & 0.0334 & 4.8940 & 0.0116 & 0.0342 & 4.8906 & 0.0115 & 0.0342\\
AlphaGAN  & 5.6696 & 0.0119 & 0.0408 & 5.2367 & 0.0112 & 0.037 & 5.2669 & 0.0112 & 0.0373\\
GCA & 4.4002 & 0.0089 & 0.0310  & 4.3469 & 0.0089 & 0.0306  & 4.3593 & 0.0088 & 0.0307 \\
IndexNet & 5.8509 & 0.0204 & 0.0422  & 5.2188 & 0.0158 & 0.0370& 5.1959 & 0.0156 & 0.0368 \\
FBA & 4.1544 & 0.0086 & 0.0290 & 4.1267 & 0.0088 & 0.0289  & 4.1330 & 0.0088 & 0.0289 \\
 \hline
\end{tabular}}
\end{center}
\caption{Results of trimap-based deep learning methods on P3M-500-P. ``B'' denotes the blurred images while ``N'' denotes the normal images. ``B:N'' denotes training on blurred images while testing on normal images, vice versa.
}
\label{tab:benchmark_trimap_based_dl}
\end{table*}

\begin{table*}[htb]
\begin{center}
\resizebox{0.9\textwidth}{!}{
\begin{tabular}{l|ccc|ccc|ccc}
\hline
Setting & \multicolumn{3}{c}{N:N} & \multicolumn{3}{|c}{B:N} & \multicolumn{3}{|c}{B:B} \\
\hline
Method & SAD & MSE  & MAD & SAD & MSE & MAD & SAD & MSE & MAD\\
\hline
SHM & 17.13 & 0.0075& 0.0099 & 24.33& 0.0116& 0.0140 & 21.56 & 0.0100 & 0.0125\\
LF & 31.22 &0.0123& 0.0181 & 30.84& 0.0129& 0.0178  & 42.95 & 0.0191 & 0.0250\\
HATT &22.93  &0.0040 &0.0133&26.5 &0.0055 &0.0155 &25.99  &0.0054  &0.0152 \\
MODNet &13.11 &0.0038 &0.0076 &13.80&0.0041 &0.0080&13.31& 0.0038& 0.0077\\
GFM &10.73& 0.0033& 0.0063 & 13.08& 0.0050& 0.0080 & 13.20 & 0.0050 & 0.0080 \\
 \hline
 BASIC & 15.52& 0.0060& 0.0090 & 14.52 &0.0054 &0.0085 & 15.13 & 0.0058 & 0.0088\\
 P3M-Net (R34) & 8.73 & 0.0026 & 0.0051 & 9.22& 0.0028& 0.0053& 9.06 &0.0028 &0.0053 \\
 P3M-Net (Swin-T) &7.19 &0.0021  &0.0042  &7.30 &0.0021 &0.0042 &7.13 &0.0021 &0.0042 \\
 P3M-Net (ViTAE-S) &6.60 &0.0018 &0.0038  &6.31 &0.0015 &0.0037 &6.24 &0.0015 &0.0036 \\
 \hline
\end{tabular}}
\end{center}
\caption{Results of trimap-free methods on P3M-500-P. Please refer to Table~\ref{tab:benchmark_trimap_based_dl} for the meaning of different symbols.}
\label{tab:benchmark_e2e}
\end{table*}

\subsection{Benchmark Setup}

We evaluate both trimap-based and trimap-free matting methods including traditional and deep learning ones on P3M-10k. The full list of the methods are shown in Table~\ref{tab:benchmark_trimap_based_traditional},~\ref{tab:benchmark_trimap_based_dl},~\ref{tab:benchmark_e2e}. We use the common metrics including MSE, SAD, and MAD to evaluate their performance. For trimap-based methods, the metrics are only calculated over the transition area, while for trimap-free methods, they are calculated over the whole image.

To evaluate methods' generalization ability under PPT setting, we train and evaluate them under three protocols: 1) trained on blurred images, tested on blurred ones (B:B); 2) trained on blurred images and tested on normal ones (B:N); and 3) trained on normal images and tested on normal ones (N:N). All the evaluation are conducted on P3M-500-P validation set for a fair comparison, with or without privacy content preserved.

\subsection{Study on the Impact of PPT}
\subsubsection{Impact on Trimap-based Traditional Methods}

We benchmarked multiple trimap-based traditional methods including Closed~\citep{levin2007closed}, IFM~\citep{aksoy2017designing}, KNN~\citep{chen2013knn}, Compre~\citep{shahrian2013improving}, Robust~\citep{wang2007optimized}, Learning~\citep{zheng2009learning}, Global~\citep{he2011global} and Shared~\citep{gastal2010shared} in Table~\ref{tab:benchmark_trimap_based_traditional}. As seen from the table, trimap-based traditional methods show neglectable performance variance under different training and evaluation protocols, indicating that the PPT setting brings little impact on these methods. This observation is reasonable, since traditional methods mainly make prediction based on local pixels in the transition area, where no blurring occurs, although a few of sampled neighboring pixels may be blurred. Note that we define the transition area as in previous works but exclude the blurred area.

\subsubsection{Impact on Trimap-based Deep Learning Methods}

We also benchmarked many trimap-based deep learning methods including DIM~\citep{dim}, AlphaGAN~\citep{bmvcLutzAS18}, GCA~\citep{li2020natural}, IndexNet~\citep{lu2019indices} and FBA~\citep{forte2020fbamatting} in the Table~\ref{tab:benchmark_trimap_based_dl}. As shown in the table, similar to traditional trimap-based methods, deep learning methods also show very minor changes across different settings. This is because trimap-based deep learning methods use the ground truth trimap as an auxiliary input and focus on estimating the alpha matte of the transition area, probably guiding the model to pay less attention to the blurred areas. In addition, there are also some observations opposed to intuition. When testing on normal images, models trained on the normal training images surprisingly fall behind of those trained on the blurred ones. For instance, the SAD of IndexNet on ``N:N'' is 0.6 higher than the score on ``B:N''. Similar results can also be found for AlphaGAN and GCA in Table~\ref{tab:benchmark_trimap_based_dl}.
We suspect that the blurred pixels near the transition area may serve as a random noise during the training process, which makes the model more robust and leads to a better generalization performance. 

\begin{figure*}[t]
    \centering
    \includegraphics[width = \linewidth]{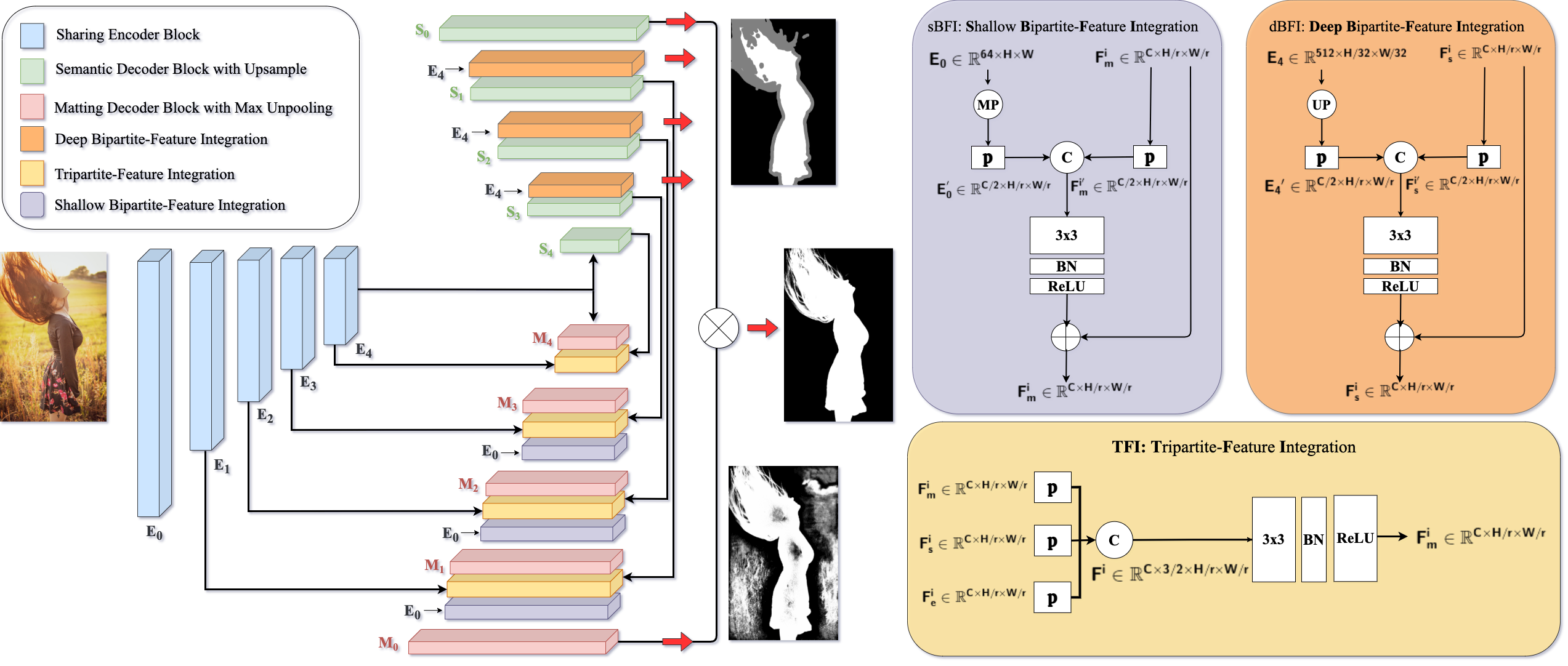}
    \caption{Diagram of the proposed P3M-Net structure. It adopts a multi-task framework, which consists of a sharing encoder, a segmentation decoder, and a matting decoder. Specifically, a TFI module, a dBFI module, and a sBFI module are devised to model different interactions among the encoder and the two decoders. Red arrows denote the network's outputs.}
    \label{fig:network}
\end{figure*}

\subsubsection{Impact on Trimap-free Methods}

Different from trimap-based methods, trimap-free methods show significant performance changes under three protocols. We benchmarked several methods including SHM~\citep{shm}, LF~\citep{lf}, MODNet~\citep{modnet}, HATT~\citep{hatt} and GFM~\citep{gfm} on P3M-10k. We summarize the results in Table~\ref{tab:benchmark_e2e} and gain some interesting insights about their generalization abilities under the PPT setting.

\begin{figure*}[!t]
\centering
\captionsetup[subfloat]{labelformat=empty,justification=centering}
\subfloat[(a) Multi-task structure]{\includegraphics[width=.48\linewidth]{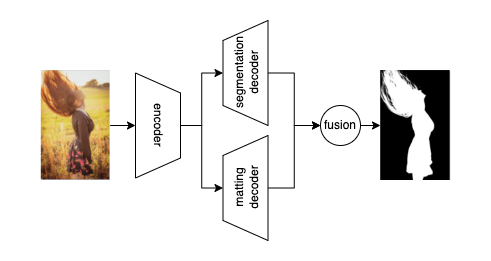}}
\subfloat[(b) Two-stage structure]{\includegraphics[width=.48\linewidth]{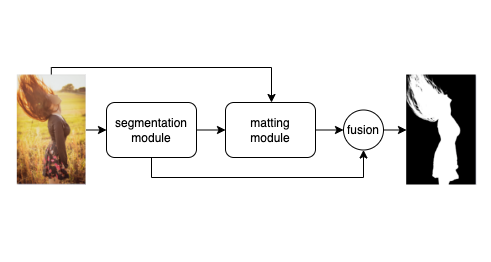}}
\caption{Diagram of the model structures of ``multi-task'' methods and ``two-stage'' methods.}
\label{fig:multi_task_and_two_stage_structures_comparison}
\end{figure*}

\textbf{First}, we start with evaluating models' generalization abilities and the impact of PPT training by comparing the results on the ``B:N'' and ``N:N'' settings. Models trained on normal training images (N:N) usually outperform those using the blurred ones (B:N), from 24.33 to 17.13 in SAD for SHM~\citep{shm}. This observation makes sense since there is a domain gap between blurred images and normal ones due to face obfuscation. By comparison, we found that trimap-free methods show different abilities in dealing with this domain gap. For example, SHM has the largest drop of 7 SAD, while MODNet~\citep{modnet} and GFM~\citep{gfm} only show a drop less than 3 in SAD. We suspect that an end-to-end multi-task framework like GFM and MODNet, which share a common encoder to learn visual features, and two task-specific decoders for segmentation and matting, perform better than others due to their joint optimization nature. By contrast, a two-stage method like SHM~\citep{shm}, which tackle the problem by two sequential stages as a segmentation stage and an auxiliary input-based matting stage, may produce segmentation errors, which can mislead the following matting stage and is difficult to be corrected. A visual comparison of `multi-task' and `two-stage' structures is provided in Figure~\ref{fig:multi_task_and_two_stage_structures_comparison}. To validate this hypothesis, we devise a baseline model called ``BASIC'' by adopting a similar multi-task framework like GFM and MODNet but removing the bells and whistles, i.e., only using a sharing encoder and two individual decoders. As shown in Table~\ref{tab:benchmark_e2e}, the small performance drop (less than 1 in SAD) proves its superiority in overcoming domain gap and supports our hypothesis.

\textbf{Second}, we further evaluate their performance in the ``B:B'' setting. The SAD scores vary from 42.95 for LF~\citep{lf} to 11.37 for MODNet~\citep{modnet}. Similarly, the models based on a unified framework like GFM, MODNet and BASIC still perform better than others with sequential matting or two-stage structure, validating the superiority of joint optimization of both sub-tasks.

These results suggest that it is better to develop a matting model based on the unified multi-task framework for P3M, which probably has a good performance on face-blurred images as well as generalizes well on normal images.

\label{sec:ppt}

%%%%%%%%%%%%%%%%%%%%%%%%%
%%% Section4. P3M-Net
%%%%%%%%%%%%%%%%%%%%%%%%%

\section{A Strong Baseline for P3M: P3M-Net}\label{sec:p3mnet}

\subsection{P3M-Net Framework}

As discussed in Section~\ref{sec:ppt}, trimap-free matting models benefit from explicitly modeling both semantic segmentation and detail matting tasks and jointly optimizing them in an end-to-end multi-task framework. Therefore, we follow GFM~\citep{gfm} to adopt the multi-task framework, where base visual features are learned from a sharing encoder and task-relevant features are learned from individual decoders, i.e., semantic decoder and matting decoder, respectively. Both decoders have five blocks, each with three convolution layers. Different upsampling operations are used in each task, i.e., bilinear interpolation in semantic decoder for simplicity and max unpooling with indices in matting decoder to preserve fine details.

Most of the previous matting methods either model the interaction between encoder and decoder such as the U-Net~\citep{unet} style structure in \citep{gfm} or model the interaction between two decoders like the attention module in~\citep{hatt}. In this paper, we try to model the comprehensive interactions between the sharing encoder and two decoders through three carefully designed integration modules, i.e., 1) a tripartite-feature integration (TFI) module to enable the interaction between encoder and two decoders; 2) a deep bipartite-feature integration (dBFI) module to enhance the interaction between the encoder and segmentation decoder; and 3) a shallow bipartite-feature integration (sBFI) module to promote the interaction between the encoder and matting decoder.

\subsubsection{TFI: Tripartite-Feature Integration}
\label{sec:tim}
Specifically, for each \textbf{TFI}, it has three inputs, i.e., the feature map of the previous matting decoder block $\mathbf{F_m^i} \in \mathbb{R}^{C\times H/r \times W/r}$, the feature map from same level semantic decoder block $\mathbf{F_s^i} \in \mathbb{R}^{C\times H/r \times W/r}$, and the feature map from the symmetrical encoder block $\mathbf{F_e^i} \in \mathbb{R}^{C\times H/r \times W/r}$, where $i\in \{1,2,3,4\}$ stands for the block index, $r$ stands for the downsample ratio of the feature map compared to the input size, and $r=2^i$. For each feature map, we use an $1\times1$ convolutional projection layer $\mathcal{P}$ for further embedding and channel reduction. The output of $\mathcal{P}$ for each feature map is $\mathbf{F^i} \in \mathbb{R}^{C/2\times H/r \times W/r}$. We then concatenate the three embedded feature maps and feed them into a convolutional block $\mathcal{C}$ containing a
$3\times3$ convolutional layer, a batch normalization layer, and a ReLU layer. As shown in Eq.~\ref{equa:tim}, the output feature is $\mathbf{F_m^i}\in\mathbb{R}^{C\times H/r \times W/r}$:
\begin{equation}
\mathbf{F_m^i} = \mathcal{C}(Concat(\mathcal{P}(\mathbf{F^i_m}),\mathcal{P}(\mathbf{F^i_s}),\mathcal{P}(\mathbf{F^i_e}))).
\label{equa:tim}
\end{equation}

\begin{figure*}[t]
    \centering
    \includegraphics[width = \linewidth]{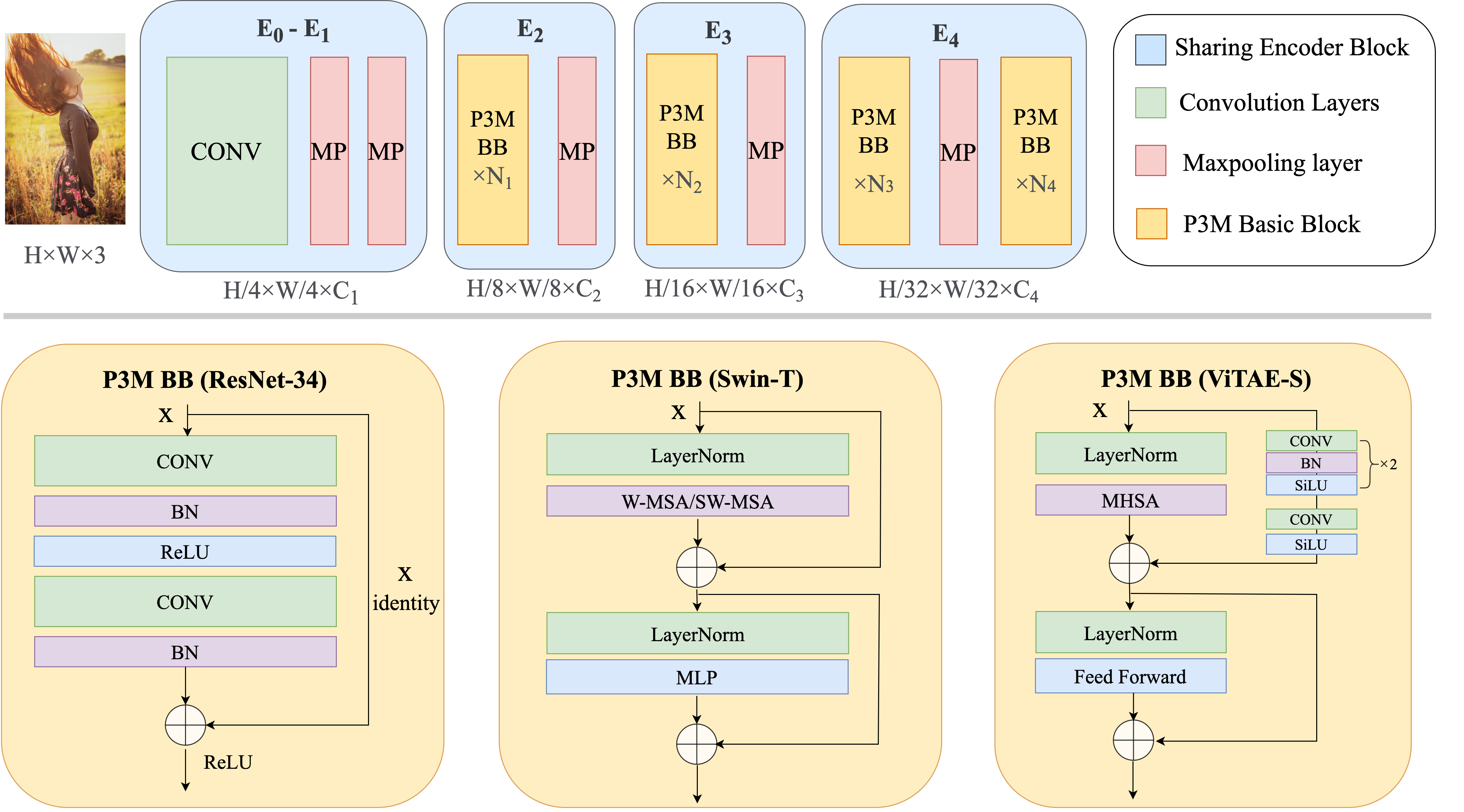}
    \caption{Illustration of the sharing encoder structure and different variants of the P3M basic blocks.}
    \label{fig:encoder}
\end{figure*}

\begin{figure*}[!t]
    \centering
    \includegraphics[width = \linewidth]{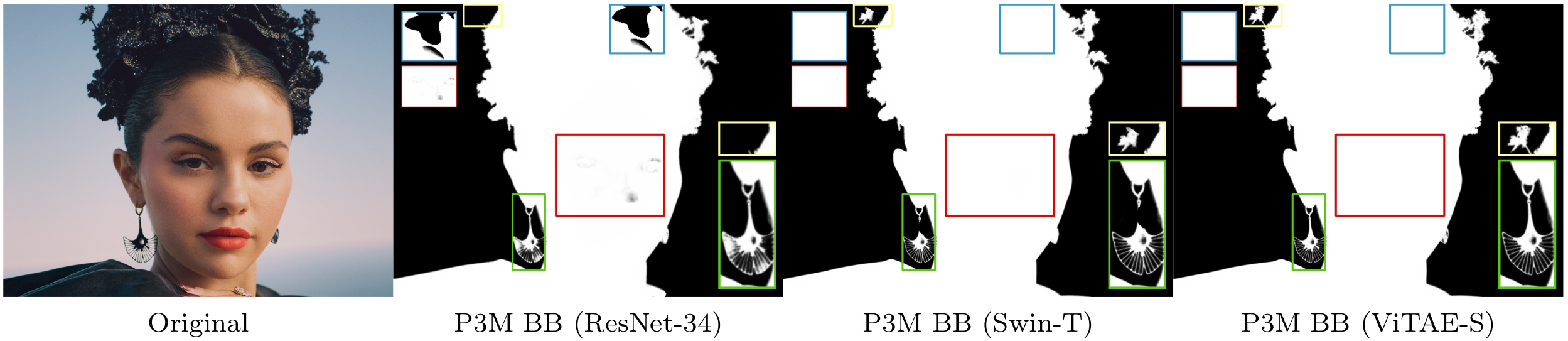}
    \caption{Visual results of different P3M variants on a test image. Closed-up views are shown in the corner.}
    \label{fig:variants_compare}
\end{figure*}

\subsubsection{sBFI: Shallow Bipartite-Feature Integration}
With the assumption that shallow layers in the encoder contain abundant structural detail features and are useful to provide matting decoder with fine foreground details, we propose the shallow bipartite-feature integration (sBFI) module. Specifically, sBFI takes the feature map $\mathbf{E_0}\in \mathbb{R}^{64\times H \times W}$ in the first encoder block as a guidance to refine the feature map $\mathbf{F_m^i}\in\mathbb{R}^{C\times H/r \times W/r}$ from the previous matting decoder block. Here, $i\in \{1,2,3\}$ stands for the layer index, $r$ stands for the downsample ratio of the feature map compared to the input size, and $r=2^i$. Since $\mathbf{E_0}$ and $\mathbf{F_m^i}$ are with different resolution, we first adopt max pooling $MP$ with a ratio $r$ on $\mathbf{E_0}$ to generate a low-resolution feature map $\mathbf{E_0^{'}}\in \mathbb{R}^{64\times H/r \times W/r}$. We then feed both $\mathbf{E_0^{'}}$ and $\mathbf{F_m^i}$ to two projection layers $\mathcal{P}$ implemented by $1\times1$ convolution layers for further embedding and channel reduction, i.e., from $C$ to $C/2$. Finally, the two feature maps are concatenated and and fed into a convolutional block $\mathcal{C}$ containing a $3\times3$ convolutional layer, a bacth normalization layer, and a ReLu layer. As shown in Eq.~\ref{equa:sgr}, we adopt the residual learning idea by adding the output feature map back to the input matting decoder feature map $\mathbf{F_m^i}$. In this way, sBFI helps the matting decoder block to focus on the fine details guided by $\mathbf{E_0}$.
\begin{equation}
\mathbf{F_m^i} = \mathcal{C}(Concat(\mathcal{P}(\mathcal{MP}(\mathbf{E_0})),\mathcal{P}(\mathbf{F^i_m})))+\mathbf{F_m^i}.
\label{equa:sgr}
\end{equation}

\subsubsection{dBFI: Deep Bipartite-Feature Integration}
Same as sBFI, features in the encoder can also provide valuable guidance to the segmentation decoder. In contrast to sBFI, we chose the feature map $\mathbf{E_4}\in \mathbb{R}^{512\times H/32 \times W/32}$ from the last encoder block, since it encodes abundant global semantics. Specifically, we devise the deep bipartite-feature integration (dBFI) module to fuse it with the feature map $\mathbf{F_s^i}\in\mathbb{R}^{C\times H/r \times W/r}$ from the $i$th segmentation decoder block to improve the feature representation ability for the high-level semantic segmentation task. Here, $i\in \{1,2,3\}$. Note that since $\mathbf{E_4}$ is in low-resolution, we use a upsampling operation $UP$ with a ratio $32/r$ on $\mathbf{E_4}$ to generate $\mathbf{E_4^{'}}\in \mathbb{R}^{512\times H/r \times W/r}$. we then feed both $\mathbf{E_4^{'}}$ and $\mathbf{F_s^i}$ into two projection layers $\mathcal{P}$, concatenated together, and fed into a convolutional block $\mathcal{C}$. We adopt the identical structures for $\mathcal{P}$ and $\mathcal{C}$ as those in sBFI. Similarly, this process can be described as follows. Note that we reuse the symbols of $\mathcal{C}$ and $\mathcal{P}$ in Eq.~\ref{equa:tim}, Eq.~\ref{equa:sgr}, and Eq.~\ref{equa:dgr} for simplicity, although each of them denotes a specific layer (block) in TFI, sBFI, and dFI, respectively.
\begin{equation}
\mathbf{F_s^i} = \mathcal{C}(Concat(\mathcal{P}(\mathcal{UP}(\mathbf{E_4})),\mathcal{P}(\mathbf{F^i_s})))+\mathbf{F_s^i}.
\label{equa:dgr}
\end{equation}

\subsection{P3M-Net Variants}
\label{sec:variants}
By integrating the above three modules into the P3M-Net framework, we are able to utilize the visual features learned from the sharing encoder comprehensively and exchange the features at global and local levels promptly. However, how to improve the representation ability of the sharing encoder to make it feasible for portrait matting remains as a challenge. As shown in Figure~\ref{fig:encoder}, we design five blocks $\mathbf{E_0}$ to $\mathbf{E_4}$ in the encoder, adopt maxpooling layers instead of strided convolution in the sharing encoder to reserve the details in the encoder as much as possible. Specifically, in $\mathbf{E_0}$ and $\mathbf{E_1}$, there are several convolution layers and maxpooling layers involved. From $\mathbf{E_2}$ to $\mathbf{E_4}$, each block has a series of P3M Basic Blocks (P3M BB) and one maxpooling layer, while $\mathbf{E_4}$ has an extra series of P3M Basic Blocks.

In the following part, we design multiple variants of P3M Basic Blocks based on CNN and vision transformers. We leverage the ability of transformers in modeling long-range dependency to extract more accurate global information and the locality modelling ability to reserve lots of details in the transition areas, and expect they can deliver better performance on both face-blurred and non-privacy images under the PPT setting. The structures of these variants are illustrated in Figure~\ref{fig:encoder} and the details are presented as follows. 

\subsubsection{P3M BB (ResNet-34)}
Following previous matting methods that adopt CNN basic block in their structures, e.g. AIM~\citep{aim}, HATT~\citep{hatt}, SHM~\citep{shm}, we use the residual block in ResNet-34~\citep{he2016deep} as our basic version of P3M BB. In particular, it consists of two convolution layers followed by a BN and ReLU. The residual structure is used to ease the training process. The number of the basic blocks in the sharing encoder from $\mathbf{N_1}$ to $\mathbf{N_4}$ are 3, 4, 6, 3 by following the original design of ResNet-34. The numbers of output channels from $\mathbf{C_1}$ to $\mathbf{C_4}$ in this case are 256, 128, 64 and 64.

Based on the above P3M BB, we obtain the ``P3M-Net (ResNet-34)'' variant, which is able to extract the contour of the human and achieves good results on most cases, as demonstrated in Figure~\ref{fig:variants_compare}. Nevertheless, there are still room for further improvement in regards to both accurate semantic information and precise details when facing the complex case. The shortage of using CNN-only basic block is twofold: 1) it is still sensitive to blurred artefacts, resulting in wrong prediction around face region as shown in the red box in the figure; and 2) the limited ability of CNN in modeling long-range dependency leads to error semantics in the alpha matte prediction as shown in the blue box in the figure. 

\subsubsection{P3M BB (Swin-T)}

Based on the above analysis, it is necessary to model the long-range dependency among pixels to enhance the model's ability in perceiving different semantic content. A reasonable solution is to adopt transformer-based basic block in our sharing encoder. As shown in Figure~\ref{fig:encoder}, we leverage the Swin~\citep{swin} transformer layers as the P3M BB, which consists of a shifted window based multi-head self attention (MSA) module with a 2-layer MLP with GELU non-linearity in between. Compared with the original full attention mechanism, the window based MSA is more computationally efficient while maintaining the ability of modelling long-range dependency. The number of the basic blocks used in the sharing encoder from $\mathbf{N_1}$ to $\mathbf{N_4}$ are 2, 2, 6, 2 by following the original design of Swin-T. The numbers of output channels from $\mathbf{C_1}$ to $\mathbf{C_4}$ in this case are 384, 192, 96 and 96.

Based on the Swin-based P3M BB, we obtain the ``P3M-Net (Swin-T)'' variant. As shown in Figure~\ref{fig:variants_compare}, it has better ability in understanding the foreground semantics, i.e., 1) becomes more insensitive to blurring artefacts as shown in the red box; and 2) provides correct semantic results compared with its CNN counterpart as shown in the blue box. Nevertheless, the predicted alpha matte of the transition area as enclosed by the green and yellow boxes misses some details. We suspect it is caused by the lack of the locality modelling ability of convolutions.

\subsubsection{P3M BB (ViTAE-S)}

Based on the above observation and analysis, we find that it is important to take the advantage of transformer-based structure for modelling long-range dependency and CNN-based structure for modelling locality, which coincide with the inherent requirement of the semantic segmentation task and detail matting task in trimap-free matting. To this end, we adopt the NC block proposed in ViATE~\citep{vitae} as our P3M BB. It consists of two parallel branches responsible for modeling long-range dependency and locality respectively, followed by a feed forward network for feature transformation. Specifically, the transformer branch consists of a multi-head self-attention module, while the local branch consists of two groups of convolution, BN, SiLU layers followed by another one convolution layer and SiLU layer. The number of the basic blocks used in the sharing encoder from $\mathbf{N_1}$ to $\mathbf{N_4}$ are 2, 2, 12, 2 as the original design of ViTAE-S. The numbers of output channels from $\mathbf{C_1}$ to $\mathbf{C_4}$ are 256, 128, 64 and 64.

Based on the ViTAE-based P3M BB, we obtain the ``P3M-Net (ViTAE-S)'' variant. As can be seen from Figure~\ref{fig:variants_compare}, compared with the CNN-only and Transformer-only basic blocks, the ViTAE-based one is able to predict correct foreground as well as fine details in the transition area simultaneously. As shown in the yellow and green boxes, the fine details of small but meticulous objects like the earring and hairpin can be distinguished clearly in the predicted alpha matte. More results of the three variants are shown in Section~\ref{sec:exp}.

%%%%%%%%%%%%%%%%%%%%%%%%%
%%% Section5. P3M-CP
%%%%%%%%%%%%%%%%%%%%%%%%%

\section{A Simple yet Effective Data and Feature Augmentation Strategy for P3M: P3M-CP}\label{sec:p3mcp}

\subsection{Overview of P3M-CP}

\begin{figure}
    \centering
    \includegraphics[width=\linewidth]{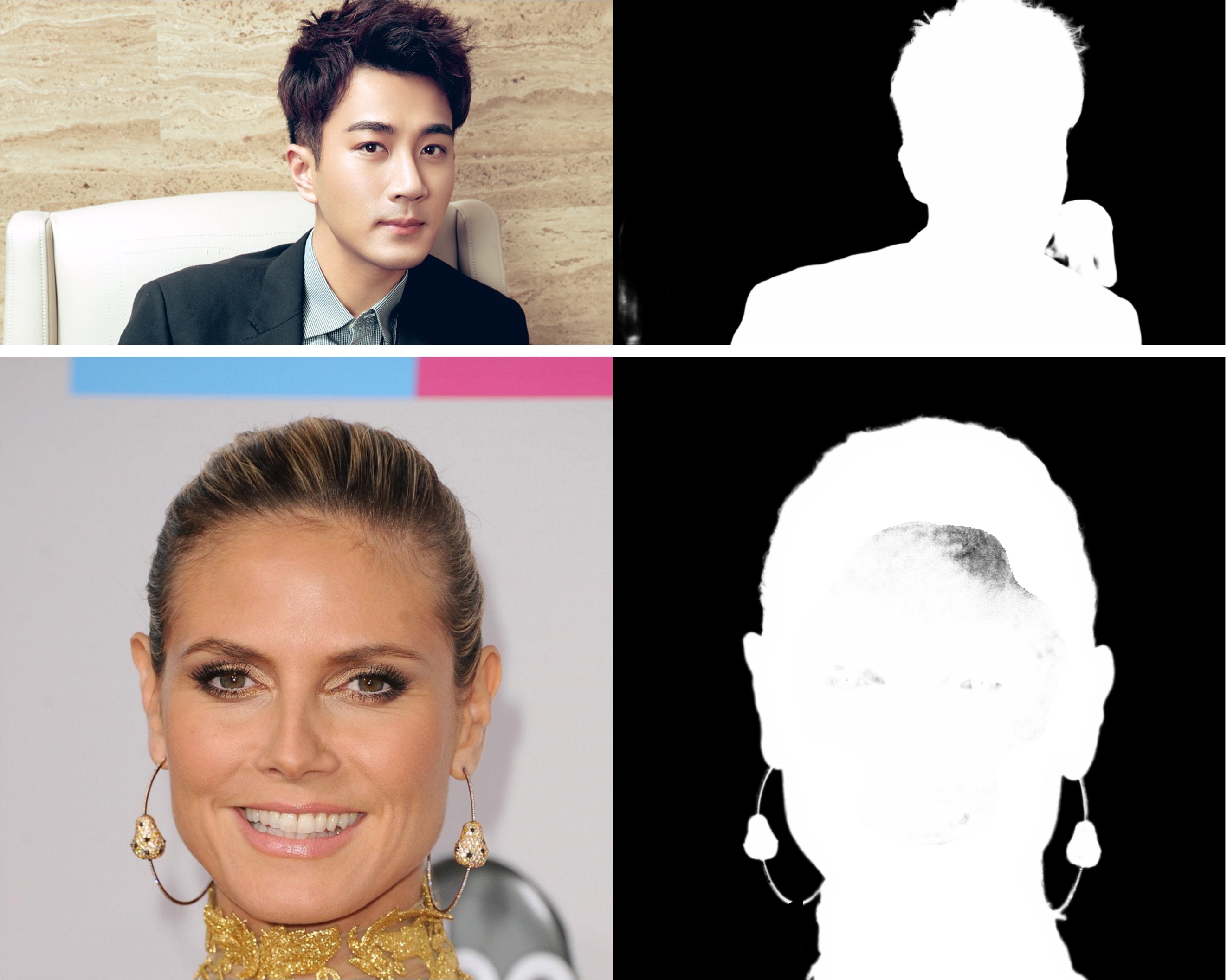}
    \caption{Some failure results of MODNet (Top) and P3M-Net (Swin-T) (Bottom) under the PPT setting on non-privacy images.}
    \label{fig:p3m-cp-motivation}
\end{figure}

\begin{figure*}
    \centering
    \includegraphics[width=\linewidth]{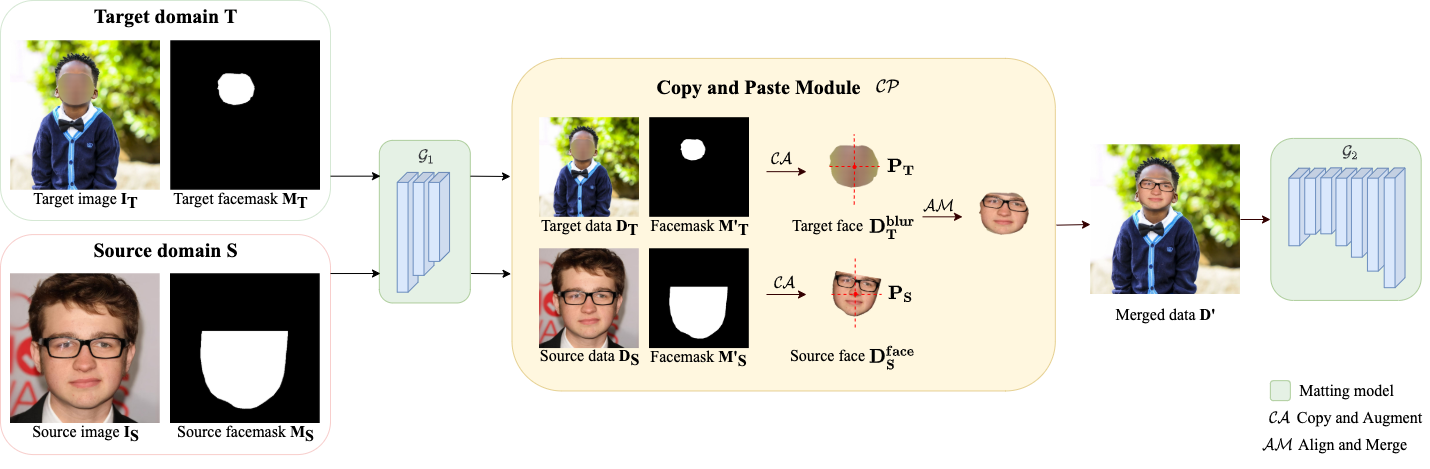}
    \caption{The pipeline of P3M-CP. P3M-CP can be applied at the image level or the feature level (as described above).}
    \label{fig:p3m-cp}
\end{figure*}

Although the P3M-Net model variants achieve better performance than previous methods as shown in the bottom rows in Table~\ref{tab:benchmark_e2e} and Figure~\ref{fig:variants_compare}, there is still a performance gap when testing on face-blurred images and generalizing on normal non-privacy images, especially for CNN-based P3M-Net. How to compensate for the lack of facial details in face-blurred training data to reduce the performance gap remains unresolved. Without seeing faces during training, the matting model is less effective in discriminating them from background, thereby affecting its generalization ability on non-privacy images. Some failure examples are shown in Figure~\ref{fig:p3m-cp-motivation}. 

To mitigate this problem, we devise a simple yet effective Copy and Paste strategy (P3M-CP) that can borrow facial information from publicly available celebrity images without privacy concerns and guide the network to reacquire the face context at both data and feature level. As shown in Figure~\ref{fig:p3m-cp}, P3M-CP adopts a Copy and Paste Module $\mathcal{CP}$ to process the images (or the features of images generated by part of the matting model, i.e., $\mathcal{G}_1$) from the source domain $\mathbf{S}$ and target domain $\mathbf{T}$ and generate the merged data $\mathbf{D'}$, which is fed into the complete matting model $\mathcal{G}_1$ and $\mathcal{G}_2$ (or the remaining part of the matting model, i.e., $\mathcal{G}_2$).

The source domain $\mathbf{S}$ consists of external facial images or portrait images $\mathbf{I_S}$ without privacy concerns, along with their facemasks $\mathbf{M_S}$ generated in advance. The target domain $\mathbf{T}$ contains the face-blurred images $\mathbf{I_T}$ from the training set of P3M-10k, along with their facemasks $\mathbf{M_T}$ that indicate which part is obfuscated to protect privacy. The copy and paste module $\mathcal{CP}$ takes both the source domain data and target domain data as inputs, copy the face region from the source data and then paste it onto the target data. In this module, the source and target data can be either images or the features extracted by part of matting model ($\mathcal{G}_1$), resulting in P3M-ICP at the image level and P3M-FCP at the feature level.

\textbf{Copy and Paste Module}. As shown in the yellow box in Figure~\ref{fig:p3m-cp}, the copy and paste module $\mathcal{CP}$ consists of two steps, (1) $\mathcal{CA}$: copy and augment, and (2) $\mathcal{AM}$: align and merge. No extra learnable weights are required during this process. First, $\mathcal{CA}$ takes the source data $\mathbf{D_S}$ (images or features) and its facemask $\mathbf{M'_S}$ as input. The original facemask is resized to the same size as the source data, denoted $\mathbf{M'_S}$. Then, the face area $\mathbf{D_S^{face}}$ in the source data $\mathbf{D_S}$ is cut out based on the facemask $\mathbf{M_S^{'}}$ and augmented by random resize and rotation. Similarly, the face-blurred area $\mathbf{D_T^{blur}}$ is cropped out from the target data $\mathbf{D_T}$ based on the resized facemask $\mathbf{M'_T}$ without augmentation. Second, the augmented source face $\mathbf{D_S^{face}}$ and the target face-blurred part $\mathbf{D_T^{blur}}$ are further processed by $\mathcal{AM}$. Specifically, the center points $\mathbf{P_S}$ and $\mathbf{P_T}$ of $\mathbf{D_S^{face}}$ and $\mathbf{D_T^{blur}}$ are calculated based on their facemasks. After aligning $\mathbf{D_S^{face}}$ and $\mathbf{D_T^{blur}}$ according to the center points, we paste the overlap region from $\mathbf{D_S^{face}}$ to $\mathbf{D_T^{blur}}$, we get the merged data $\mathbf{D'}$. This $\mathcal{CP}$ process can be applied on both images and features, depending on the type of inputs, which can be formulated as follows:
\begin{equation}
\mathbf{D_S^{face}} = Rot(RS(\mathbf{D_S} \odot \mathbf{M'_S})),\ \mathbf{D_T^{blur}} = \mathbf{D_T} \odot \mathbf{M'_T},
\label{equa:cp-process-1}
\end{equation}
\begin{equation}
\mathbf{P_S} = center(\mathbf{D_S^{face}}), \ \mathbf{P_T} = center(\mathbf{D_T^{blur}}),
\label{equa:cp-process-2}
\end{equation}
\begin{equation}
\mathbf{D'} = \mathcal{AM}(\mathbf{D_S^{face}}, \mathbf{D_T^{blur}}, \mathbf{P_S}, \mathbf{P_T}, \mathbf{D_T}),
\label{equa:cp-process-3}
\end{equation}
where $Rot$, $RS$, $center$ denote the rotate, resize, and calculate the center operations, respectively.

\begin{figure}
    \centering
    \includegraphics[width=0.8\linewidth]{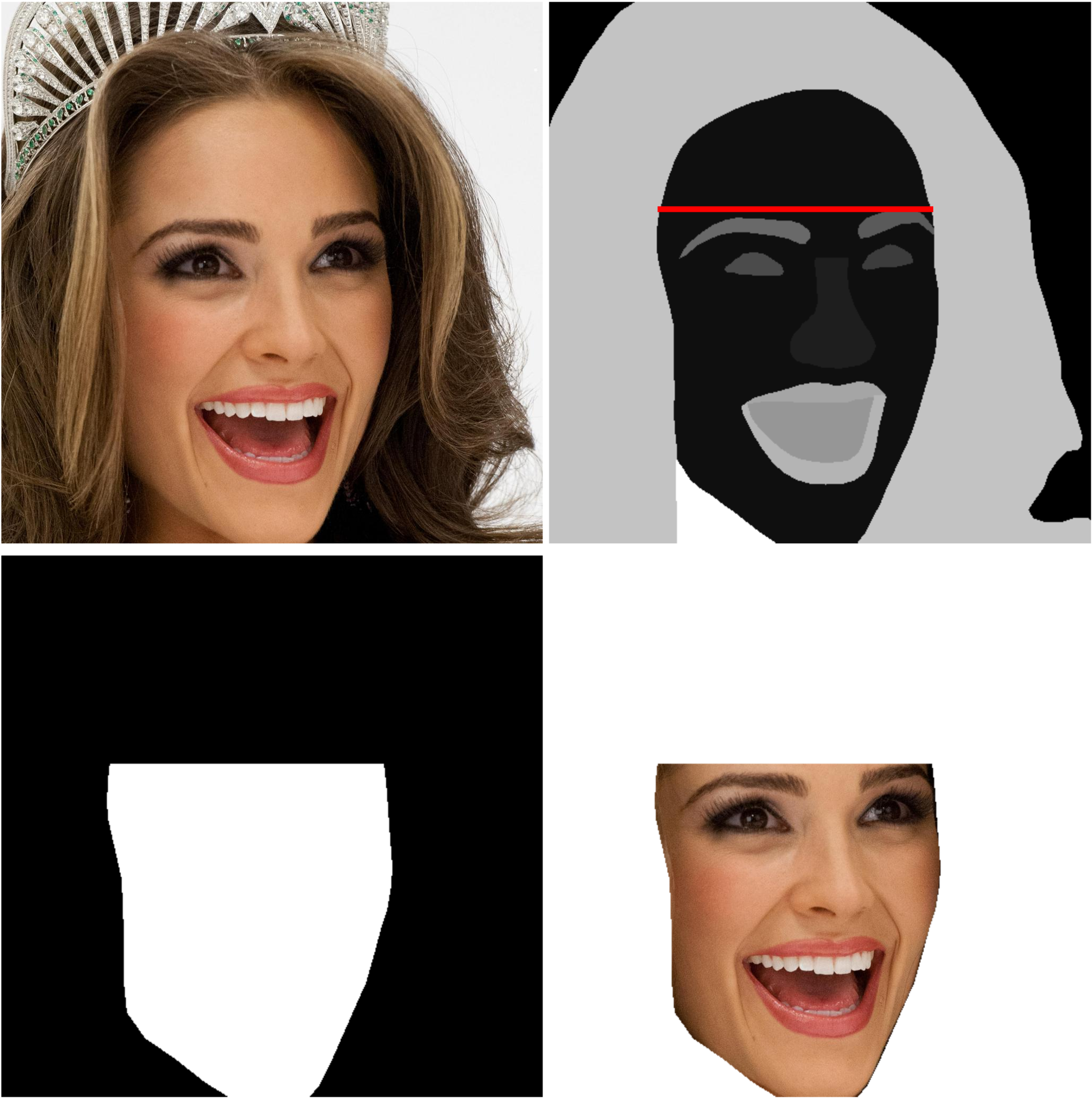}
    \caption{Illustration of the process of generating facemask for the source image. \textbf{left top}: An image from CelebAMask-HQ~\citep{CelebAMask-HQ}. \textbf{right top}: the annotation of facial parts. The red line is placed at right above the brows. \textbf{left bottom}: the generated facemask. \textbf{right bottom}: the face region.}
    \label{fig:celeba_anno}
\end{figure}

\textbf{Source Data and Facemask Annotation.} 
It is noteworthy that the source data $\mathbf{I_S}$ do not need to have alpha matte labels, which require neglectable effort to collect. Nevertheless, they should contain clear faces, which is quite difficult to collect from ordinary people in the context of privacy preserving. Luckily, we find that public celebrity images can be used for non-commercial purposes by law, which have no privacy issues. Therefore, we adopt the existing celebrity face dataset CelebAMask-HQ~\citep{CelebAMask-HQ} as the source images for P3M-CP, which also provides the annotations of facial parts. We use the annotated masks of 'skin' and 'brow' to extract the accurate face region. The process is illustrated in Figure~\ref{fig:celeba_anno}. Specifically, we first determine a line right above the left and right brows, and then take the face skin under the line as the face region.

\subsection{P3M-ICP: Copy and Paste at Image Level}
With source face images and facemasks prepared, the P3M-CP module aim to capture the facial information and use it to guide the learning process of matting models. P3M-ICP accomplishes it at image level, i.e., directly applying ``copy and paste'' process on images. The process is shown in Figure~\ref{fig:p3m-cp}, where the copy and paste module in the yellow box is applied directly on images in front of matting model $\mathcal{G}_1$. Note that in P3M-ICP, source images $\mathbf{I_S}$ and source data $\mathbf{D_S}$ are equivalent. So are the target images $\mathbf{I_T}$ and target data $\mathbf{D_T}$.

\begin{figure}
    \centering
    \includegraphics[width=\linewidth]{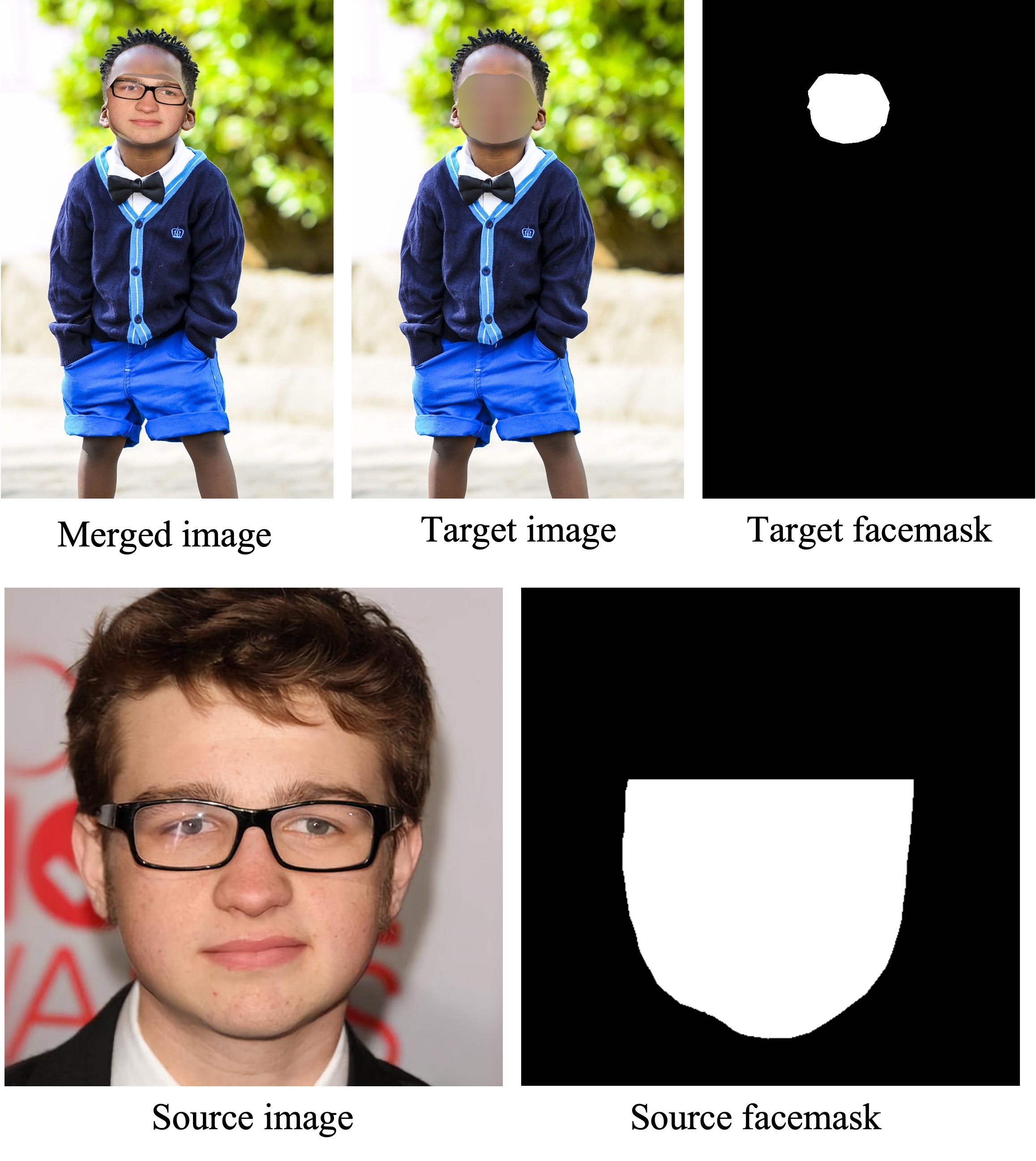}
    \caption{P3M-ICP examples. Top: merged image after P3M-ICP, source face-blurred image, and source facemask. Bottom: source image and source facemask.}
    \label{fig:data-cp-examples}
\end{figure}

When training with P3M-ICP, we randomly select a pair of images from CelebAMask-HQ dataset and the training set, and apply P3M-ICP on them with a probability of 0.5. Some examples are provided in Figure~\ref{fig:data-cp-examples}. P3M-ICP is an easy-to-implement plug-in module without extra learnable parameters, which can serve as a flexible data augmentation method and is compatible with any matting model. Besides, it only brings a few additional computations during training, while enabling the matting model to process both face-blurred images and the non-privacy ones pretty well without extra effort during inference.

Although P3M-ICP shows superior generalization ability on non-privacy images (see Section~\ref{sec:p3mcp-results}), it can be improved in the future in many aspects. For example, the face-context may be inconsistent between the source and target images, regarding the size, pose, emotion, gender of the face. More research efforts are needed to match those attributes of source and target images.

\subsection{P3M-FCP: Copy and Paste at Feature Level}
Directly cropping the face region out from the source image neglects the face context information, which may result in an incomplete face representation. To address this issue, we propose P3M-FCP that captures and merges face and context information at feature level. As shown in Figure~\ref{fig:p3m-cp}, instead of straight removing the context residing in source image and only pasting the face area onto the target image, P3M-FCP first extract features $\mathbf{D_S}$ and $\mathbf{D_T}$ of both source image $\mathbf{I_S}$ and target face-blurred image $\mathbf{I_T}$ using the first few layers in the encoder of the matting model, denoted as $\mathcal{G}_1$. In this way, the face context information in the source image could be propagated to and embedded in the face area. Then, $\mathcal{CP}$ is conducted on the features $\mathbf{D_S}$ and $\mathbf{D_T}$, along with their resized facemasks $\mathbf{M'_S}$ and $\mathbf{M'_T}$, same as that in P3M-ICP. Finally, we get the merged feature $\mathbf{D'}$ from the P3M-FCP module and feed it to the remaining part of the matting model, denoted as $\mathcal{G}_2$. Similar to P3M-ICP, this process can be formulated as follows:
\begin{equation}
    \mathbf{D_S} = \mathcal{G}_1(\mathbf{I_S}), \mathbf{D_T} = \mathcal{G}_1(\mathbf{I_T}),
\label{equa:p3m-fcp-1}
\end{equation}
\begin{equation}
   \mathbf{D'} = \mathcal{CP}(\mathbf{D_S}, \mathbf{M'_S}, \mathbf{D_T}, \mathbf{M'_T}),
\label{equa:p3m-fcp-2}
\end{equation}
where $\mathcal{CP}$ is defined as in Eq.~\ref{equa:cp-process-1} $\sim$ Eq.~\ref{equa:cp-process-3}.

In implementation, we pass both the source image $\mathbf{I_S}$ and the target training image $\mathbf{I_T}$ through the encoder of the matting model, and only apply the P3M-FCP module on a single selected layer in the encoder with a probability of 0.5. 
Usually, using P3M-FCP on a shallow layer could deliver better results. After P3M-FCP, the merged features go through the remaining part of the matting model, while the features of the source image are abandoned. Besides, the gradients corresponding to the the source image would not be back-propagated. At each iteration, only one source image is selected for each mini-batch of training images. The batch size is usually set to 8.

Compared to P3M-ICP, the P3M-FCP module can extract more face context information residing in source images and merge the source and the target features in a more consistent manner through end-to-end training. Therefore, with less demand on source data, P3M-FCP can achieve better generalization performance on non-privacy images (see Section~\ref{sec:p3mcp-results}).

%%%%%%%%%%%%%%%%%%%%%%%%%
%%% Section6. Experiment
%%%%%%%%%%%%%%%%%%%%%%%%%

\begin{table*}[htb]
\begin{center}
\resizebox{\linewidth}{!}{
\begin{tabular}{c|c|ccccc|c|cccc}
\hline
 \multicolumn{2}{c|}{Method} & LF & HATT & SHM & MODNet & GFM & DIM$\star$ & P3M(R) & P3M(S) & P3M(S)+FCP & P3M(V)\\
 \multicolumn{2}{c|}{Backbone} & DenseNet-201 & ResNeXt & PSPNet-50+VGG16 & MobileNetV2 & ResNet-34 & VGG16 & ResNet-34 & Swin-T & Swin-T & ViTAE-S \\
\hline
&SAD & 42.95 & 25.99 & 21.56 & 13.31 & 13.20  & - & 8.73 & 7.13 & 7.43 & 6.24 \\
&MSE & 0.0191 & 0.0054 & 0.0100 & 0.0038 &0.0050  & - & 0.0026 & 0.0021 & 0.0022 & 0.0015 \\
&MAD & 0.0250 & 0.0152& 0.0125& 0.0077 &0.0080& - &0.0051 & 0.0042 & 0.0043 & 0.0036 \\
&SAD-T &12.43 &11.03 & 9.14&8.11& 8.84& 4.89& 6.89& 5.82 & 5.80 & 5.65 \\
P3M-500-P&MSE-T &0.0421 &0.0377 &0.0255&0.0258 &0.0269  &0.0115 &0.0193 & 0.0153 & 0.0151 & 0.0142 \\
&MAD-T &0.0824 &0.0752 &0.0545 &0.0563&0.0616 &0.0342 &0.0478 & 0.0398 & 0.0395 & 0.0385 \\
&SAD-FG &18.922 & 2.575&0.486 &2.777 & 0.872& - & 0.673& 0.447 & 0.563 & 0.184 \\
&SAD-BG & 11.595 &12.385 &13.098 &2.424 & 3.487& - &1.166 & 0.867 & 1.075 & 0.409 \\
&Grad & 42.19& 14.91& 21.24&16.50&12.58& 4.48& 8.22 & 11.94 & 11.59 & 10.94 \\
&Conn & 18.80& 25.29& 17.53&10.88& 17.75 &9.68&13.68 & 6.73 & 7.03 & 5.86 \\
\hline
&SAD & 32.59& 30.53& 20.77&16.70 &15.50& - &11.23 & 8.89 & 7.94 & 7.59 \\
&MSE &0.0131 &0.0072 &0.0093 &0.0051&0.0056 & - &0.0035& 0.0026 & 0.0021 & 0.0019 \\
&MAD &0.0188 &0.0176 &0.0122 &0.0097&0.0091& - &0.0065 & 0.0052 & 0.0047 & 0.0044 \\
&SAD-T & 14.53& 13.48& 9.14& 9.13&10.16 & 5.32& 7.65& 6.73 & 6.51 & 6.60 \\
P3M-500-NP&MSE-T & 0.0420& 0.0403& 0.0255&0.0237& 0.0268 & 0.0094& 0.0173& 0.0149 & 0.0138 & 0.0142 \\
&MAD-T &0.0825 &0.0803 &0.0545 &0.0549&0.0620&0.0324 &0.0466 & 0.0401 & 0.0390 & 0.0396 \\
&SAD-FG &8.924 &2.930 &0.935 & 3.143&2.172 & - & 1.414& 1.253 & 0.439 & 0.148 \\
&SAD-BG & 9.136&14.121 &10.701 &4.434&3.161  & - &2.165 & 0.909 & 0.988 & 0.838 \\
&Grad & 31.93& 19.88& 20.30&15.29 &14.82 &4.70 &10.35 & 10.79 & 10.33 & 9.84 \\
&Conn & 19.50& 27.42& 17.09&13.81& 18.03& 7.70& 12.51& 8.18 & 7.25 & 6.96 \\
\hline
\end{tabular}}
\end{center}
\caption{Results of the three P3M-Net variants and some representative methods on P3M-500-P and P3M-500-NP. DIM$\star$ uses ground truth trimaps. The backbones of each method from left to right are, DenseNet-201~\citep{huang2017densely}, ResNeXt~\citep{resnext}, PSPNet-50~\citep{zhao2017pyramid}+VGG16~\citep{vgg16}, MobileNetV2~\citep{mobilenetv2}, ResNet-34~\citep{he2016deep}, VGG16~\citep{vgg16}, ResNet-34~\citep{he2016deep}, Swin-T~\citep{swin}, Swin-T~\citep{swin}, and ViTAE-S~\citep{vitae}, respectively.}
\label{tab:experiment}
\end{table*}

%%%%% results on adobe portrait 636 benchmark
\begin{table*}[htb]
\begin{center}
\resizebox{\linewidth}{!}{
\begin{tabular}{c|c|ccccc|c|cccc}
\hline
\multicolumn{2}{c|}{Method} & LF & HATT & SHM & MODNet& GFM& DIM$\star$ &P3M(R)& P3M(S) & P3M(S)+FCP & P3M(V)\\
\hline
& SAD & 71.18 & 53.45 & 48.95 & 42.63 & 37.93 & - & 36.47 & 32.71 & 31.11 & 29.00 \\
& MSE & 0.0423 & 0.0234 & 0.0283 & 0.0227 & 0.0220 & - & 0.0183 & 0.0157 & 0.0149 & 0.0151 \\
& MAD & 0.0556 & 0.0376 & 0.0375 & 0.0338 & 0.0316 & - & 0.0270 & 0.0236 & 0.0231 & 0.0229 \\
& SAD-T & 31.28 & 28.99 & 25.01 & 24.84 & 25.49 & 16.98 & 22.25 & 21.38 & 21.32 & 21.43 \\
RWP test set & MSE-T & 0.1281 & 0.1170 & 0.1013 & 0.0989 & 0.0980 & 0.0508 & 0.0803 & 0.0744 & 0.0728 & 0.0760 \\
& MAD-T & 0.1972 & 0.1856 & 0.1629 & 0.1611 & 0.1672 & 0.1052 & 0.1394 & 0.1326 & 0.1325 & 0.1344 \\
& SAD-FG & 20.027 & 8.332 & 6.190 & 5.840 & 4.146 & - & 5.895 & 3.999 & 3.108 & 1.750 \\
& SAD-BG & 19.879 & 16.125 & 17.747 & 11.950 & 8.300 & - & 8.325 & 7.328 & 6.682 & 5.822 \\
& Grad & 71.64 & 78.90 & 68.52 & 66.26 & 67.56 & 46.47 & 60.82 & 57.48 & 56.83 & 56.66 \\
& Conn & 70.54 & 46.71 & 48.74 & 39.63 & 37.48 & 16.93 & 36.36 & 32.48 & 31.00 & 28.77 \\
\hline
\end{tabular}}
\end{center}
\caption{Results of the three P3M-Net variants and some representative methods on RWP test set~\citep{yu2021mask}. All the models are trained with our P3M-10k face-blurred training set. DIM$\star$ uses ground truth trimaps. P3M(R), P3M(S), and P3M(V) stand for the P3M-Net variants based on ResNet-34, Swin-T, and ViTAE-S backbones, respectively.}
\label{tab:experiment_adobe_portrait}
\end{table*}

\section{Experiments}\label{sec:experiments}
\subsection{Experiment Settings}
To compare the proposed P3M-Net with existing trimap-free methods, such as SHM~\citep{shm}, LF~\citep{lf}, HATT~\citep{hatt}, GFM~\citep{gfm}, and MODNet~\citep{modnet}, we train them on the P3M-10k face-blurred images and evaluate them on 1) the face-blurred validation set P3M-500-P, 2) the normal validation set P3M-500-NP, following the PPT setting, 3) P3M-500-P, processed by different privacy-preserving methods, e.g., blurring, mosaicing, and masking, and 4) RWP test set~\citep{yu2021mask} consisting of 636 normal portrait images for testing. Furthermore, we apply P3M-ICP and P3M-FCP on MODNet~\citep{modnet} and the P3M-Net variants during training, and validate them on P3M-500-P and P3M-500-NP.

\noindent\textbf{Implementation Details}
For training P3M-Net, we crop a patch from the image with a size randomly chosen from $512\times512$, $768\times768$, $1024\times1024$, and then resize it to $512\times512$. We randomly flip the patches for data augmentation. The learning rate is fixed as $1\times10^{-5}$. We train P3M-Net variants on NVIDIA Tesla V100 GPUs with a batch size of 8 for 150 epochs, which takes about 2 days. It takes 0.132s to test on an $800\times800$ image. For GFM~\citep{gfm}, LF~\citep{lf} and MODNet~\citep{modnet}, we use the code provided by the authors. For SHM~\citep{shm}, HATT~\citep{hatt} and DIM~\citep{dim} which have no official code, we re-implement them. For P3M-ICP or P3M-FCP, we apply them on each image with a probability of 0.5 during training.

\noindent\textbf{Evaluation Metrics} We follow previous works and adopt the evaluation metrics including the sum of absolute differences (SAD), mean squared error (MSE), mean absolute difference (MAD), gradient (Grad.) and Connectivity (Conn.)~\citep{rhemann2009perceptually}. We calculated them over the whole image for trimap-free methods. We also report the SAD-T, MSE-T, MAD-T metrics to calculate the score within the transition area, and SAD-FG, SAD-BG to calculate the score within the foreground and background, respectively.

\subsection{Results and Analysis of P3M-Net Variants}
\subsubsection{Objective and Subjective Results}
\label{sec:exp}

\begin{figure*}[hbtp]
    \centering
    \includegraphics[width = 0.95\linewidth]{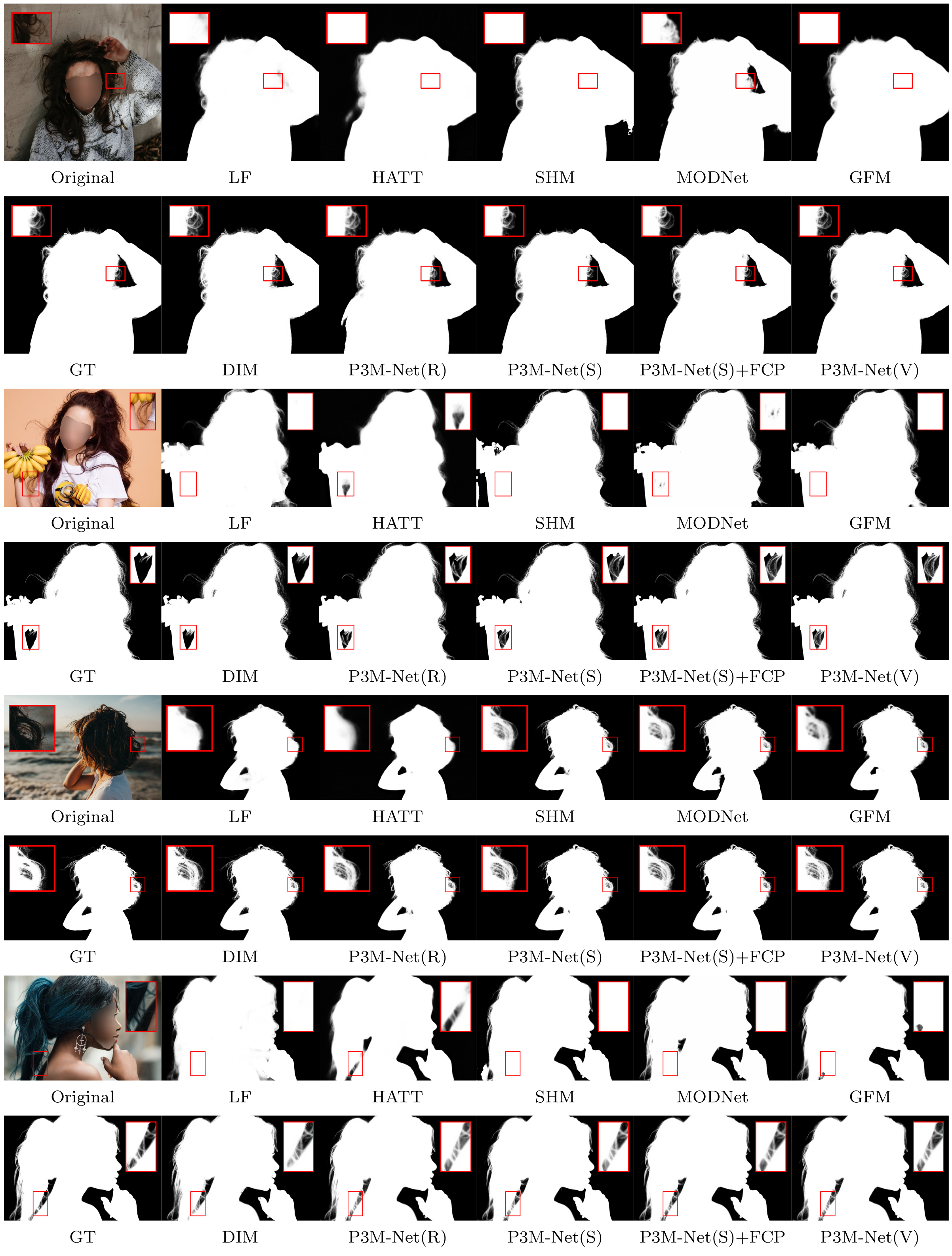}
    \caption{Visual results of SOTA methods and the proposed P3M-Net variants on P3M-500-P. Among all the methods, only DIM~\citep{dim} requires an extra trimap as input while the others are automatic methods.}
    \label{fig:variants_compare_results_val500p}
\end{figure*}

\begin{figure*}[hbtp]
    \centering
    \includegraphics[width = 0.94\linewidth]{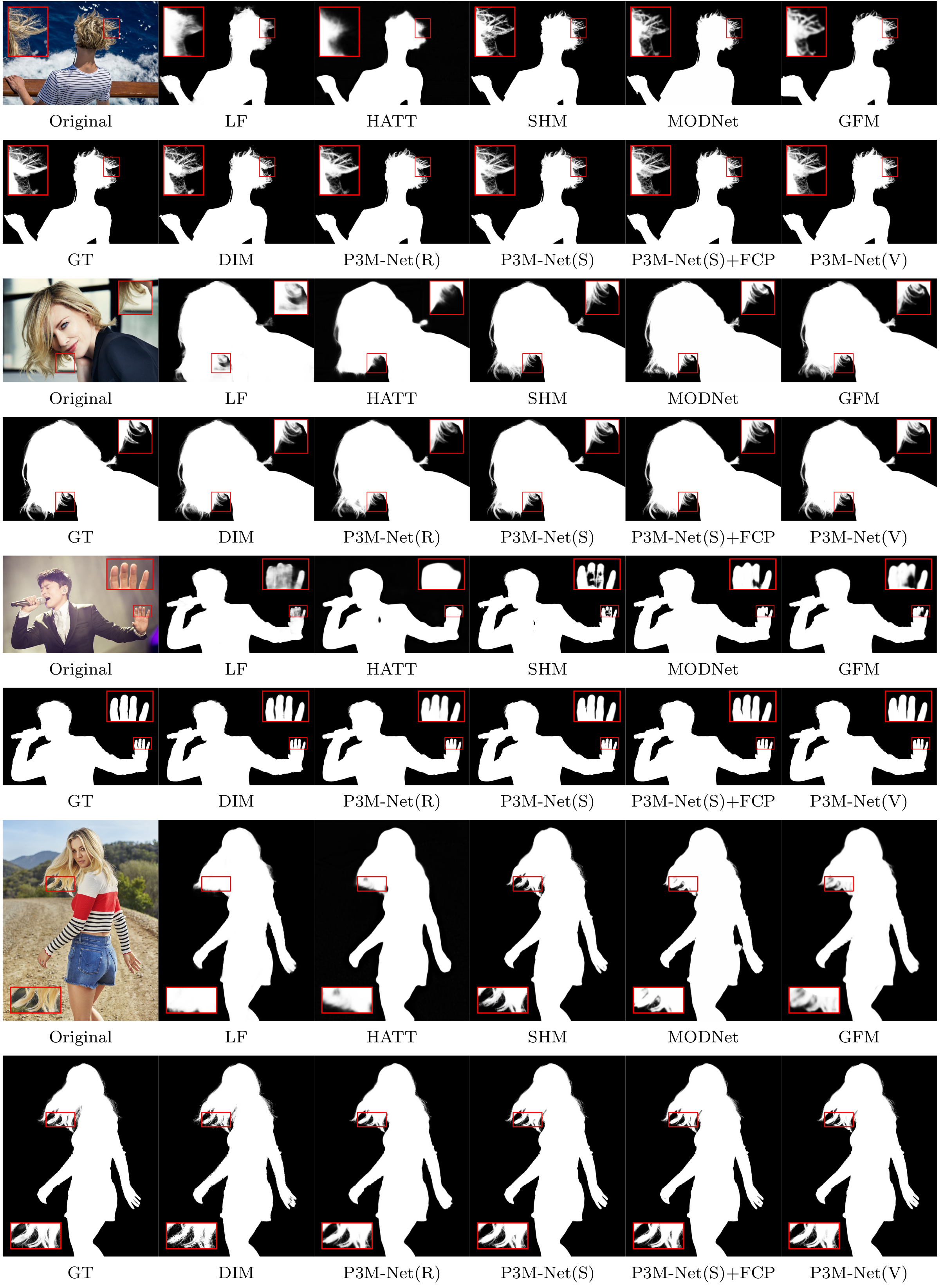}
    \caption{Visual results of SOTA methods and the proposed P3M-Net variants on P3M-500-NP. Among all the methods, only DIM~\citep{dim} requires an extra trimap as input while the others are automatic methods.}
    \label{fig:variants_compare_results_val500np}
\end{figure*}

The objective and subjective results of the proposed P3M-Net variants and some representative methods are shown in Table~\ref{tab:experiment}, Figure~\ref{fig:variants_compare_results_val500p}, and Figure~\ref{fig:variants_compare_results_val500np}, respectively. As can be seen, all variants of P3M-Net outperform the previous trimap-free methods in all metrics and even achieve competitive results with trimap-based method DIM~\citep{dim}, which requires the ground truth trimap as an auxiliary input, denoted as DIM$\star$. These results validate the design of the three integration modules which are able to model abundant interactions between encoder and decoder as well as the three P3M BB variants which leverage the advantages of long-range dependency modelling ability and locality modelling ability. As for SHM~\citep{shm}, it has worse SAD than all P3M-Net variants on both validation sets, i.e., 21.56 v.s. 6.24 and 20.77 v.s. 7.59, due to its two-stage pipeline, which produces many segmentation errors that is difficult to be corrected in the following stage. LF~\citep{lf} and HATT~\citep{hatt} have large error in transition areas, e.g., 12.43 and 11.03 SAD v.s. 5.65 SAD of ours, since they lack explicit semantic guidance for the matting task. As in Figure~\ref{fig:variants_compare_results_val500p}, they have ambiguous segmentation results and inaccurate matting details. MODNet~\citep{modnet} and GFM~\citep{gfm} are able to predict more accurate foreground and background owing to the multi-task learning framework. However, they may fail to predict correct context and have worse performance than ours, i.e., 13.32 and 13.20 v.s. 6.24 in SAD, since it lacks of exploring feature interactions between encoder and decoders. DIM~\citep{dim} has lower SAD than ours since it uses ground truth trimap. Nevertheless, all P3M-Net variants still achieve competitive performance in the transition area, e.g., 6.89 v.s. 4.89 SAD.

Although all three P3M-Net variants have already surpasses the SOTA methods, they show different performance due to the use of different backbones. As discussed in Section~\ref{sec:variants}, comparing with P3M-Net (ResNet-34), P3M-Net (Swin-T) reduces the foreground and background errors from 3.579 to 2.162 on P3M-500-NP, owing to its long-range dependency modelling ability for better semantic perception. The errors have been decreased to 1.427 when utilizing P3M-FCP on P3M-Net (Swin-T) further. When adopting the ViTAE P3M BB, it can be further reduced to 0.986, a dramatic drop by \textbf{73\%} compared with the P3M-Net (ResNet-34). On the other hand, P3M-Net (Swin-T) has already reduce the SAD error in the transition area from 6.89 to 5.82 and 7.65 to 6.73 on P3M-500-P and P3M-500-NP, respectively. The SAD error can be further reduced to 5.65 and 6.60 by P3M-Net (ViTAE-S). These results confirm the superiority of the parallel structure in ViTAE-S that can extract useful global and local features for semantic segmentation and detail matting.

\begin{figure*}[hbtp]
    \centering
    \includegraphics[width = 1\linewidth]{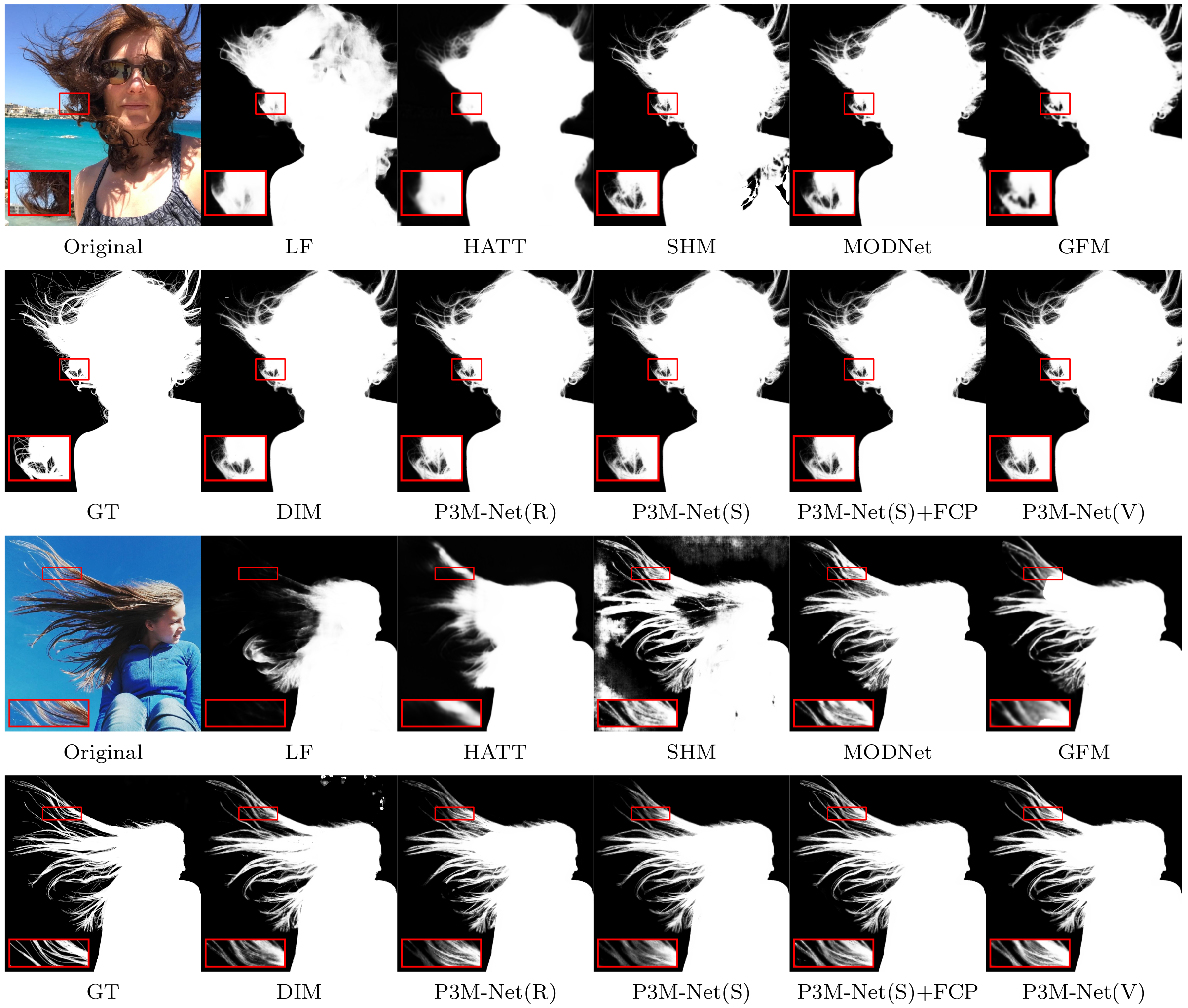}
    \caption{Visual results of SOTA methods and the proposed P3M-Net variants on RWP test set~\citep{yu2021mask}. Among all the methods, only DIM~\citep{dim} requires an extra trimap as input while the others are automatic methods.}
    \label{fig:variants_compare_results_realworld}
\end{figure*}

\subsubsection{Ablation Study}
We conduct ablation study of P3M-Net on P3M-500-P and P3M-500-NP validation sets. As can be seen from Table~\ref{tab:ablation}, the basic multi-task baseline without any our proposed modules can achieve a fairly good result compared with previous methods \citep{shm,hatt,lf}. With TFI, SAD decreases dramatically to 11.32 and 13.7, owing to the valuable semantic features from encoder and segmentation decoder for matting. Besides, sBFI (dBFI) decreases SAD from 11.32 to 9.47 (9.76) on P3M-500-P and from 13.7 to 12.36 (12.45) on P3M-500-NP, confirming their values in providing useful guidance from relevant visual features. With all three modules, the SAD decreases from 15.13 to 8.73, and 17.01 to 11.23, indicating that our proposed modules bring about 50\% relative performance improvement. We also count the model parameters (million) of each ablation set for a fair comparison. As seen from the table, from the basic version to P3M-Net with all three proposed modules, the model parameters increase only 1.81M but make a large improvement on the performance.

\begin{table*}[htb]
\begin{center}
\resizebox{\linewidth}{!}{
\begin{tabular}{c|ccc|c|ccc|ccc}
\hline
\multicolumn{5}{c}{}  &\multicolumn{3}{|c}{P3M-500-P} & \multicolumn{3}{|c}{P3M-500-NP} \\
\hline
Model & TFI & sBFI & dBFI & \#Params (M) & SAD & MSE & MAD & SAD & MSE & MAD\\
\hline
 BASIC & &  &  & 37.67 &15.13 & 0.0058 & 0.0088 & 17.01 &0.0062  &0.0099\\
  - & $\checkmark$& &   & 38.93 & 11.32 & 0.0042 & 0.00066 & 13.70 & 0.0052 &0.008 \\
 - & $\checkmark$& $\checkmark$ &  & 39.17 &9.47  & 0.0030 & 0.0055 &12.36  &0.0043  &0.0072  \\
 - & $\checkmark$&  & $\checkmark$  & 39.24 &9.76  &0.0031  & 0.0057 & 12.45 & 0.0043 &0.0073 \\
 \hline
 P3M-Net & $\checkmark$&$\checkmark$  &$\checkmark$  & 39.48 & \textbf{8.73} &\textbf{0.0026}  &\textbf{0.0051}  & \textbf{11.23} & \textbf{0.0035} & \textbf{0.0065}\\
 \hline
\end{tabular}}
\end{center}
\caption{Ablation study of the key modules in P3M-Net. The BASIC version stands for the bare multi-task framework without any addition modules.}
\label{tab:ablation}
\end{table*}

\begin{table}[htb]
\begin{center}
% \resizebox{\linewidth}{!}{
\begin{tabular}{c|c|ccc}
\hline
& & SAD & MSE & MAD \\
\hline
& blurring & 8.73 & 0.0026 & 0.0051 \\
P3M-Net(R) & mosaicing &8.55 & 0.0028 & 0.0050 \\
& masking &8.99 & 0.0030 & 0.0052 \\
\hline
& blurring & 7.13 & 0.0021 & 0.0042 \\
P3M-Net(S) & mosaicing & 7.01 & 0.0020 & 0.0041 \\
& masking & 7.71 & 0.0024 & 0.0045 \\
\hline
& blurring & 6.24 & 0.0015 & 0.0036 \\
P3M-Net(V) & mosaicing & 6.50 & 0.0017 & 0.0038 \\
& masking & 6.56 & 0.0017 & 0.0038 \\
\hline
\end{tabular}
\end{center}
\caption{Results of P3M-Net variants on P3M-500-P, where the test images are processed by different privacy-preserving methods, including blurring, mosaicing, and masking. P3M-Net(R), P3M-Net(S), and P3M-Net(V) stand for the P3M-Net variants based on the ResNet-34, Swin-T, and ViTAE-S backbones, respectively.}
\label{tab:experiment-p3m-different-obfuscation}
\end{table}

\begin{table*}[htb]
\begin{center}
\resizebox{\linewidth}{!}{
\begin{tabular}{c|c|cccc|cccc|cccc}
\hline
\multicolumn{2}{c|}{Method} & \multicolumn{4}{c|}{P3M-Net(R)} & \multicolumn{4}{c|}{P3M-Net(S)} & \multicolumn{4}{c}{MODNet} \\
\hline
\multicolumn{2}{c|}{CP Module} & - & +ICP & +FCP & $\dag$ & - & +ICP & +FCP & $\dag$ & - & +ICP & +FCP & $\dag$ \\
\hline
& SAD & 8.73 & 8.41 & 8.13 & - & 7.13 & 7.11 & 7.43 & - & 13.31 & 12.66 & 12.92 & - \\
& MSE & 0.0026 & 0.0027 & 0.0026 & - & 0.0021 & 0.0020 & 0.0022 & - & 0.0038 & 0.0035 & 0.0037 & - \\
& MAD & 0.0051 & 0.0049 & 0.0047 & - & 0.0042 & 0.0041 & 0.0043 & - & 0.0077 & 0.0074 & 0.0075 & - \\
& SAD-T & 6.89 & 6.26 & 6.24 & - & 5.82 & 5.79 & 5.80 & - & 8.11 & 7.91 & 7.94 & - \\
P3M-500-P & MSE-T & 0.0193 & 0.0178 & 0.0181 & - & 0.0153 & 0.0152 & 0.0151 & - & 0.0258 & 0.0248 & 0.0253 & - \\
& MAD-T & 0.0478 & 0.0426 & 0.0430 & - & 0.0398 & 0.0397 & 0.0395 & - & 0.0563 & 0.0552 & 0.0552 & - \\
& SAD-FG & 0.673 & 0.695 & 0.592 & - & 0.447 & 0.449 & 0.563 & - & 2.777 & 3.096 & 3.404 & - \\
& SAD-BG & 1.166 & 1.460 & 1.293 & - & 0.867 & 0.871 & 1.075 & - & 2.424 & 1.658 & 1.579 & - \\
\hline
& SAD & 11.23 & 9.13 & 9.07 & 9.11 & 8.89 & 8.21 & 7.94 & 7.99 & 16.70 & 14.69 & 14.69 & 14.54 \\
& MSE & 0.0035 & 0.0027 & 0.0026 & 0.0028 & 0.0026 & 0.0023 & 0.0021 & 0.0021 & 0.0051 & 0.0040 & 0.0041 & 0.0040 \\
& MAD & 0.0065 & 0.0053 & 0.0052 & 0.0054 & 0.0052 & 0.0048 & 0.0047 & 0.0047 & 0.0097 & 0.0086 & 0.0085 & 0.0084 \\
& SAD-T & 7.65 & 6.84 & 6.76 & 7.07 & 6.73 & 6.48 & 6.51 & 6.64 & 9.13 & 8.76 & 8.55 & 8.43 \\
P3M-500-NP & MSE-T & 0.0173 & 0.0156 & 0.0152 & 0.0168 & 0.0149 & 0.0138 & 0.0138 & 0.0142 & 0.0237 & 0.0220 & 0.0217 & 0.0207 \\
& MAD-T & 0.0466 & 0.0409 & 0.0409 & 0.0425 & 0.0401 & 0.0388 & 0.0390 & 0.0399 & 0.0549 & 0.0530 & 0.0520 & 0.0511 \\
& SAD-FG & 1.414 & 0.863 & 1.190 & 0.731 & 1.253 & 1.005 & 0.439 & 0.452 & 3.143 & 3.757 & 3.753 & 3.931 \\
& SAD-BG & 2.165 & 1.426 & 1.124 & 1.317 & 0.909 & 0.727 & 0.988 & 0.898 & 4.434 & 2.177 & 2.388 & 2.178 \\
\hline
\end{tabular}}
\end{center}
\caption{Results of different models using P3M-CP on P3M-500-P and P3M-500-NP. ``-'' means the model is trained on the face-blurred training set without using any P3M-CP strategy. $\dag$ means the model is trained on the normal training set. P3M-Net(R) and P3M-Net(S) stand for the P3M-Net variants based on the ResNet-34 and Swin-T backbones, respectively.}
\label{tab:experiment-p3m-cp}
\end{table*}

\begin{table}[htb]
\begin{center}
\resizebox{\linewidth}{!}{
\begin{tabular}{c|c|cccc}
\hline
\multicolumn{2}{c|}{Method} & \multicolumn{4}{c}{P3M-Net(V)} \\
\hline
\multicolumn{2}{c|}{CP Module} & - & +ICP & +FCP & $\dag$\\
\hline
& SAD & 6.24 & 6.85 & 6.58 & - \\
& MSE & 0.0015 & 0.0019 & 0.0017 & - \\
& MAD & 0.0036 & 0.0040 & 0.0038 & - \\
& SAD-T & 5.65 & 5.80 & 5.84 & - \\
P3M-500-P & MSE-T & 0.0142 & 0.0149 & 0.0153 & - \\
& MAD-T & 0.0385 & 0.0396 & 0.0400 & - \\
& SAD-FG & 0.184 & 0.780 & 0.312 & - \\
& SAD-BG & 0.409 & 0.269 & 0.433 & - \\
\hline
& SAD & 7.59 & 7.62 & 7.54 & 7.39 \\
& MSE & 0.0019 & 0.0018 & 0.0018 & 0.0017 \\
& MAD & 0.0044 & 0.0044 & 0.0044 & 0.0043 \\
& SAD-T & 6.60 & 6.58 & 6.78 & 6.59 \\
P3M-500-NP & MSE-T & 0.0142 & 0.0138 & 0.0144 & 0.0136 \\
& MAD-T & 0.0396 & 0.0395 & 0.0406 & 0.0395 \\
& SAD-FG & 0.148 & 0.566 & 0.138 & 0.153 \\
& SAD-BG & 0.838 & 0.473 & 0.622 & 0.634 \\
\hline
\end{tabular}}
\end{center}
\caption{Results of P3M-Net (ViTAE-S) (denoted as P3M-Net(V)) with P3M-CP on P3M-500-P and P3M-500-NP. ``-'' and $\dag$ have the same meanings as in Table~\ref{tab:experiment-p3m-cp}.}
\label{tab:experiment-p3m-cp-failed-case}
\end{table}

\subsubsection{Results on Different Privacy-preserved Validation Sets}
To further validate the generalization ability of P3M-Net variants on different types of privacy-preserved data, we train them with face-blurred images and evaluate them on P3M-500-P, where
the test images are processed by different privacy-preserving
methods, including blurring, mosaicing, and masking. The results are shown in Table~\ref{tab:experiment-p3m-different-obfuscation}. As can be seen, there is only a slight degradation on the test images processed by mosaicing and masking, compared to the results on blurring data. These results suggest that the matting model trained under the PPT setting can handle different types of privacy-preserved data, which is of great practical significance in real-world applications.

\begin{figure*}[hbtp]
    \centering
    \includegraphics[width=0.95\linewidth]{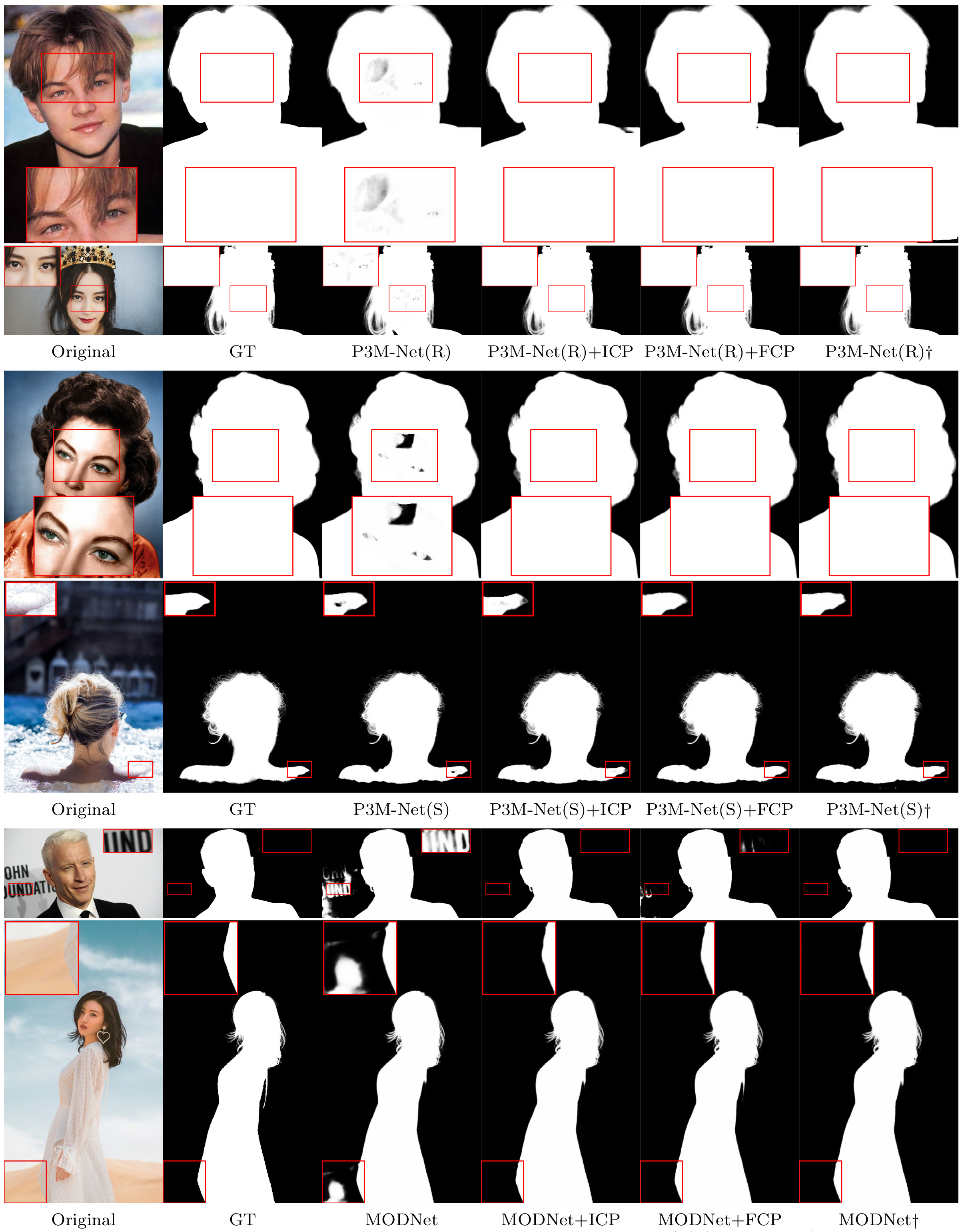}
    \caption{Visual results of P3M-CP on MODNet, P3M-Net(R), and P3M-Net(S). The test images are from P3M-500-NP. P3M-Net(R) and P3M-Net(S) stand for the P3M-Net variants based on ResNet-34 and Swin-T backbones, respectively. $\dag$ means the model is trained on the normal training set, where the real faces are available.}
    \label{fig:cp-results}
\end{figure*}

\subsubsection{Results on RWP Test Set}
To further validate the generalization ability of P3M-Net variants on normal portrait images, we train them with face-blurred images and evaluate them on RWP test set~\citep{yu2021mask}, which consists of 636 natural portrait images without privacy preserving. The objective and subjective results are listed in Table~\ref{tab:experiment_adobe_portrait} and Figure~\ref{fig:variants_compare_results_realworld}, respectively. As can be seen, all variants of P3M-Net outperform the previous trimap-free methods in almost all metrics and even achieve competitive results with trimap-based method DIM~\citep{dim}. This conclusion is similar to that shown in P3M-500-NP validation set, which further validate the superior generalization ability of our proposed P3M-Net variants. Moreover, the P3M-FCP also brings improvement on Real-world Portrait test set when applied on P3M-Net(Swin-T) model.

\subsection{Results and Analysis of P3M-CP}\label{sec:p3mcp-results}
We apply P3M-ICP and P3M-FCP on three P3M-Net variants and MODNet during training, and validate them on P3M-500-P and P3M-500-NP validation sets. The objective and subjective results are summarized in Table~\ref{tab:experiment-p3m-cp} and Table~\ref{tab:experiment-p3m-cp-failed-case}, and shown in Figure~\ref{fig:cp-results}.

\subsubsection{Experiment Settings} 
For each model, we train it under four settings and compare their generalization abilities on P3M-500-NP. The settings include 1) training on face-blurred training set, 2) training on face-blurred training set with P3M-ICP, 3) training on face-blurred training set with P3M-FCP, and 4) training on normal version of training set. Note that the images in the face-blurred training set and its normal version are the same, except that the faces in face-blurred training set are obfuscated for privacy protection. The last setting is to show the upper-bound performance we can achieve on P3M-500-NP since there is no domain gap when training on normal images and testing normal images.

\subsubsection{Objective Results} 
As can be seen in Table~\ref{tab:experiment-p3m-cp}, for each baseline, the model trained with P3M-CP achieves a significant improvement on the P3M-500-NP containing normal images in terms of all evaluation metrics, while the performance on P3M-500-P containing privacy-preserving images is almost not affected. Especially, P3M-ICP and P3M-FCP bring a large improvement by 18.7\% and 19.2\% in SAD when applied on P3M-Net (ResNet-34). Moreover, they also achieve a competitive or even better result compared with the models trained on normal training set. For example, P3M-Net (Swin-T) with P3M-FCP achieves a SAD of 7.94, smaller than that trained on normal training set, i.e., 7.99 in SAD.

Comparing the scores of SAD-T, SAD-FG, and SAD-BG, we find that the large improvement gained by P3M-CP mainly comes from the foreground and background area. Specifically, the P3M-FCP on P3M-Net (ResNet-34) reduces the SAD error by 1.265 in the foreground and background area, compared with the decrease of 0.89 in the transition area. Similar results can be observed on P3M-Net (Swin-T) and MODNet. This implies that the P3M-CP strategy can effectively compensate the absent facial context in the face-blurred training images and obtain a better semantic perception ability.

All the above results validates that our P3M-ICP and P3M-FCP can effectively reduce the domain gap and improve the generalization ability of models on non-privacy images under the PPT setting. We will discuss their impact on P3M-Net (ViTAE-S) in Sec.~\ref{subsubsec:cp-vitae}.

\subsubsection{Subjective Results} 
Figure~\ref{fig:cp-results} shows the visual results of P3M-ICP and P3M-FCP applied on the three models, i.e., P3M-Net (ResNet-34), P3M-Net (Swin-T), and MODNet. As can be seen, although P3M-Net (ResNet-34) and P3M-Net (Swin-T) achieve good objective scores on P3M-500-NP, they cannot handle the face area well. The models tend to predict the face area as background. This problem can be well resolved by using P3M-ICP and P3M-FCP during training, where reasonable prediction is obtained in the face area, similar to the models trained on normal training set. As for MODNet in Figure~\ref{fig:cp-results}, the side impact of PPT setting is mainly on the foreground and background area, where P3M-CP can also mitigate this problem effectively.

\subsubsection{Analysis of P3M-CP on P3M-Net (ViTAE-S)}
\label{subsubsec:cp-vitae}
Although P3M-CP shows significant improvement on multiple models, e.g., P3M-Net (ResNet-34), P3M-Net (Swin-T), and MODNet, it does not deliver significant improvement when applied on P3M-Net (ViTAE-S). As shown in Table~\ref{tab:experiment-p3m-cp-failed-case}, both P3M-ICP and P3M-FCP achieve similar or slightly worse results compared with the baseline without using any P3M-CP strategy. Meanwhile, the performance on face-blurred validation set P3M-500-P also degrades marginally.

These results are reasonable considering the excellent generalization ability of P3M-Net (ViTAE-S) and the limitation of P3M-CP strategies. On the one hand, according to the minor performance gap on P3M-500-NP between the model trained with face-blurred data and normal data, i.e., 7.59 v.s. 7.38 in SAD, we could believe that the domain gap has already been resolved to a large extent owing to the excellent representation ability of the parallel convolution and transformer structure in ViTAE. On the other hand, pasting the randomly aligned source faces without considering the discrepancy of age, gender, pose, emotion between the source and target faces, also introduces extra domain gap that might affect the training of P3M-Net (ViTAE-S). In other words, for models without a good generalization ability, the benefit of facial context compensation overwhelms the side impact caused by the introduced discrepancy. In contrast, for models with an excellent generalization ability like P3M-Net (ViTAE-S), the benefit is minor while introduced discrepancy may degenerate the optimization, resulting in marginal improvement or even having side impact on the models. P3M-CP could be improved in future works by matching the face attributes of source and target images.

\subsection{Model Complexity Analysis}

\begin{table}[htb]
\begin{center}
\resizebox{0.75\linewidth}{!}{
\begin{tabular}{c|cc}
\hline
Method & \#Params (M) & Inference time (s) \\
\hline
P3M(R) & 39.48 & 0.0127 \\
P3M(S) & 45.13 & 0.0228 \\
P3M(V) & 27.46 & 0.0399 \\
\hline
MODNet & 6.49 & 0.0131 \\
DIM & 130.55 & 0.0440 \\
GFM & 55.29 & 0.0135 \\
LF & 37.91& 0.1103\\
HATT &106.96 & 0.0862\\
SHM & 79.27& 0.0993\\
\hline
\end{tabular}}
\end{center}
\caption{Model complexity and inference speed comparison. In inference speed test, the input size is 512 $\times$ 512, and the inference time is averaged over 100 trials. P3M(R), P3M(S), and
P3M(V) stand for the P3M-Net variants based on ResNet-34, Swin-T, and ViTAE-S backbones, respectively.}
\label{tab:model_complexity_and_speed_sota}
\end{table}

To conduct a further analysis regards to the model complexity of our proposed P3M-Net, its variants, and the related state-of-the-art methods. We provide the number of model parameters (million, denoted as $M$) and the inference speed of each method on an image resized to $512\times512$ and show the results in Table~\ref{tab:model_complexity_and_speed_sota}. The inference speed is tested on a NVIDIA Tesla V100 GPU. As shown in the table, P3M-Net variant with ResNet-34 provides the most efficient inference time since it only takes 0.0127s to render a $512\times512$ image, which is even better than MODNet that has the smallest amount of model parameters. On the other hand, P3M-Net variant with ViTAE-S that provides the best performance compared with other state-of-the-art methods as shown in Table~\ref{tab:experiment}, has a second least parameters number as 27.46 million. The results have further validated P3M-Net's efficiency from the aspects of model parameters and inference speed.

%%%%%%%%%%%%%%%%%%%%%%%%%
%%% Section7. Conclusion
%%%%%%%%%%%%%%%%%%%%%%%%%

\section{Conclusion}\label{sec:conclusion}

In this paper, we make the first study on the privacy-preserving portrait matting (P3M) problem to respond to the increasing privacy concerns. Specifically, we define the privacy-preserving training (PPT) setting, and establish the first large-scale anonymized portrait dataset P3M-10k, containing 10,000 face-blurred images and ground truth alpha mattes. 

We empirically find that the PPT setting has little side impact on trimap-based methods while trimap-free methods perform differently, depending on their model structures. We identify that trimap-free methods using a multi-task framework that explicitly models and optimizes both segmentation and matting tasks can effectively mitigate the side impact of PPT. 

Accordingly, we provide a strong baseline model named P3M-Net, which specifically focuses on modeling the interactions between encoder and decoders, showing promising performance and outperforming all previous trimap-free methods. We further devise three variants of P3M-Net by leveraging the advantages of CNN and vision transformer backbones. Extensive experiments show all three variants outperform state-of-the-art methods. 

Furthermore, we devise a simple yet effective copy and paste strategy (P3M-CP) that can improve the generalization ability of models on non-privacy images. In the future, we will improve our P3M-CP strategy by considering the discrepancy of age, gender, pose, emotion between the source and target faces to further reduce the domain gap. We hope this study can open a new perspective for the research of portrait matting and attract more attention from the community to address the privacy concerns. 

\bibliographystyle{spbasic}      
\bibliography{matting.bib}

\end{document}